\documentclass{article}

\usepackage[final]{neurips_data_2024}

\usepackage[utf8]{inputenc} % allow utf-8 input
\usepackage[T1]{fontenc}% use 8-bit T1 fonts
\usepackage{arydshln}
\usepackage{graphicx}
\usepackage{float}
\usepackage{hyperref}       % hyperlinks
\usepackage{url}            % simple URL typesetting
\usepackage{booktabs}       % professional-quality tables
\usepackage{amsfonts}       % blackboard math symbols
\usepackage{nicefrac}       % compact symbols for 1/2, etc.
\usepackage{microtype}      % microtypography
\usepackage{xcolor}         % colors
\usepackage{array}
\usepackage{booktabs}
\usepackage{graphicx}
\usepackage{amsmath}
\usepackage{graphicx}
\usepackage{bbm}
\usepackage{bm}
\usepackage{booktabs}
\usepackage{tcolorbox}
\usepackage{amsfonts} 
\usepackage{tabularx}
\usepackage{makecell}
\usepackage{xcolor}
\usepackage{paralist}
\usepackage[capitalise,noabbrev]{cleveref}
\usepackage{graphicx}
\usepackage{wrapfig}
\usepackage{xspace}
\usepackage{bm}
\usepackage{paralist}
\usepackage{enumitem}
\usepackage{titlesec}

\crefformat{footnote}{#2\footnotemark[#1]#3}
\titlespacing{\paragraph}{0pt}{0pt}{6pt}
\setitemize{noitemsep,topsep=0pt,parsep=0pt,partopsep=0pt}

\newcommand{\ifprecedingtext}[1]{\ifvmode\relax\else#1\fi}
\renewcommand{\-}{
    \ifprecedingtext{\\}
    \textbullet\xspace
}
\renewcommand{\cite}{\citep}
\newenvironment{blue}{
    \color{blue}
}{
    \ignorespacesafterend
}
\newenvironment{orange}{
    \color{orange}
}{
    \ignorespacesafterend
}
\newenvironment{olive}{
    \color{olive}
}{
    \ignorespacesafterend
}
\newenvironment{purple}{
    \color{purple}
}{
    \ignorespacesafterend
}

\newcommand{\datasetname}{\texttt{IllusionBench}\xspace}
\newcommand{\logos}{\datasetname-\texttt{LOGO}\xspace}
\newcommand{\icons}{\datasetname-\texttt{ICON}\xspace}
\newcommand{\sindata}{\datasetname-\texttt{IN}\xspace}

\newcommand{\xc}[1]{x^C_{#1}}

\newcommand{\yc}[1]{y^C_{#1}}
\newcommand{\x}[1]{x_{#1}}
\newcommand{\tc}{\tau_{C}}
\newcommand{\ts}{\tau_{S}}
\newcommand{\tb}{\tau_{C,S}}

\title{Hidden in Plain Sight: Evaluating Abstract Shape Recognition in Vision-Language Models}

   \author{%
  Arshia Hemmat \\
  University of Oxford\\
  \And
  Adam Davies$^*$ \\
   University of Illinois Urbana-Champaign\\
     \And
  Tom A. Lamb$^*$ \\
   University of Oxford\\
  \AND
  Jianhao Yuan$^*$ \\
   University of Oxford\\
  \And
  Philip Torr \\
   University of Oxford\\
  \And
   Ashkan Khakzar\\
   University of Oxford\\
  \And
   Francesco Pinto\\
   University of Oxford\\
} 

\begin{document}

\maketitle
% Title proposals from gemini (some bits of them are not bad!)
% \begin{itemize}
% \item Beyond Recognition: A Dataset for Testing VLM Abstraction Abilities in Compositional Shape Perception
% \item Seeing Shapes in the Chaos: A Benchmark for VLM Recognition of Emergent Forms
% \item From Parts to Wholes: Unveiling VLM Capabilities in Compositional Shape Recognition
% \item Pareidolia Power: A Dataset for Evaluating VLM Abstraction from Object Compositions
% \item The Illusion of Form: Testing VLM Generalizability in Recognizing Emergent Shapes
% \item Beyond the Sum of Parts: A Dataset to Benchmark VLM Abstraction and Pareidolic Recognition
% \item Unveiling the Hidden: A Compositional Shape Dataset to Push VLM Understanding of Visual Gestalt
% \item Shape Shifters: A Dataset for Evaluating VLM Generalization in Recognizing Emergent Forms
% \item From Chaos to Coherence: A Dataset for Testing VLM Abstraction in Recognizing Compositional Shapes
% \item Seeing Through the Noise: A Benchmark for VLM Abstraction Abilities in Pareidolic Shape Perception
% \end{itemize}

\def\thefootnote{*}\footnotetext{These authors contributed equally to this work}\def\thefootnote{\arabic{footnote}}

%%%%%%%%%%%%%%%%%%%%%%%%%%%%%%%%%%%%%%%%%%%%%%%%%%%%%%%%%%%%

\begin{abstract}
% mixing all below from Ashkan, FP and Adam
Despite the importance of shape perception in human vision, early neural image classifiers relied less on shape information for object recognition than other (often spurious) features. While recent research suggests that current large Vision-Language Models (VLMs) exhibit more reliance on shape, we find them to still be seriously limited in this regard. To quantify such limitations, we introduce \datasetname, a dataset that challenges current cutting-edge VLMs to decipher shape information when the shape is represented by an arrangement of visual elements in a scene. 
Our extensive evaluations reveal that, while these shapes are easily detectable by human annotators, current VLMs struggle to recognize them, indicating important avenues for future work in developing more robust visual perception systems. The full dataset and codebase are available at: \url{https://arshiahemmat.github.io/illusionbench/}
\end{abstract}

\section{Introduction}
\label{sec:intro}

%logic: VLMs are great, can solve several tasks, an open question is whether they can abstract from some symbol and recognise it under strong semantically meaningful perturbations? we consider cases in which the object is objectively present to cases in which the object is simply hallucinated 

Deep neural networks have accomplished remarkable breakthroughs in visual recognition over the past decade ~\cite{krizhevsky2012imagenet,he2016deep,dosovitskiy2020image,radford2021learning,team2023gemini};
% Concurrent with these advancements, it soon emerged that deep vision models exhibit fundamental limits.
but these models have also shown longstanding, fundamental limitations -- for instance,
the performance of these models degrades when faced with common corruptions and perturbations ~\cite{hendrycks2019benchmarkingCommonCorrupt}, or natural out-of-distribution data ~\cite{hendrycks2021natural}.
% {\red TODO (Adam): put in summary of psych lit here (importance of shape, leads to invariance/robustness), and use this to motivate the work studying shape bias -- e.g., "given that shape is invariant and leveraging it enables robust generalization, we want to know whether/how much image classifiers use it"}
%
% What is responsible for this lack of robustness, 
How can we facilitate more robust neural vision models?
% How can we facilitate such robustness in the context of neural vision models?
A natural place to begin is by considering the source of robustness in human vision.
Human object recognition is largely based on shape perception \cite{landau1988importance,biederman1988surface,xu2004emergence,baker2018abstract}, which is essential to the robustness of human vision due to the invariance of shape to common transformations such as translation, rotation, scaling, and changes in illumination, color, and texture \cite{kendall1984shape,hummel2001complementary,ommer2013role,dryden2016statistical}.
As such, substantial work in computer vision has focused on improving and evaluating shape perception (e.g., \citealp{ritter2017cognitive,geirhos2018imagenettrained, islam2021shape,geirhos2021partial,gavrikov2024vision}, \emph{inter alia}),
finding that early deep vision models relied much more on texture than shape in image classification \cite{geirhos2018imagenettrained,islam2021shape,Pinto2022AnIT,Benarous2023HarnessingSD, subramanian2023spatialfrequency}, which is believed to contribute to their lack of robustness \cite{geirhos2020shortcut,gavrikov2024vision}. %add more, etc.
Later work observed that vision encoders trained with larger-scale data weakly supervised by language (e.g., CLIP; ~\citealp{radford2021learning}) show improvements in shape recognition ~\citep{geirhos2021partial,gavrikov2024vision}.
% Moreover, it is shown that leveraging a language module enables disentangling between shape and texture information through prompting with text. 
% and that their performance can be controlled and further enhanced via prompt engineering \cite{}.
% The finding implies that the vision module encodes and leverages both shape and texture information in classification ~\cite{gavrikov2024vision}. 
% These signify a significant advance in visual perception by deep neural networks.

While clear indicators of progress in visual perception of neural vision models,
it is important to note that all of the above studies on shape recognition in vision models have relied on two standard datasets, Cue Conflict and Stylized-ImageNet \cite{geirhos2018imagenettrained}, which presents several concerns -- for instance, these datasets do not include coherent, naturalistic visual scenes; they are built using legacy style transfer techniques that damage shape information and prevent the reproduction of fine-grained textures; and each image includes only a single object class represented as an abstract shape using perceptually uniform textures (see \cref{sec:rel_works} for a more detailed critique). %several sadamaged shape information and low-quality style transfer (as thoroughly discussed in 
% \cref{RELATEDWORKS}. %elaborated in more detail in {\red cref}):  
% these datasets are generated using outdated style transfer techniques \cite{gatys2016image,huang2017arbitrary} that are known to exhibit serious limitations \cite{wang2021evaluate}, removing key shape and scene information from original images \cite{geirhos2018imagenettrained,wang2023stylediffusion}.
%Furthermore, this benchmark was designed for closed-set image classifiers
% Even without these limitations, each image in both dataset contains only a single abstract shape corresponding to the ImageNet class of the original image represented using a perceptually uniform texture map, meaning that models are only tested in a narrow and oversimplified context.
% Thus, while these datasets have been valuable tools for assessing shape recognition in earlier models, we believe that the latest generation of VLMs merits a more sophisticated benchmark.
%
\begin{figure}[t]
    \centering
    \includegraphics[width=\textwidth]{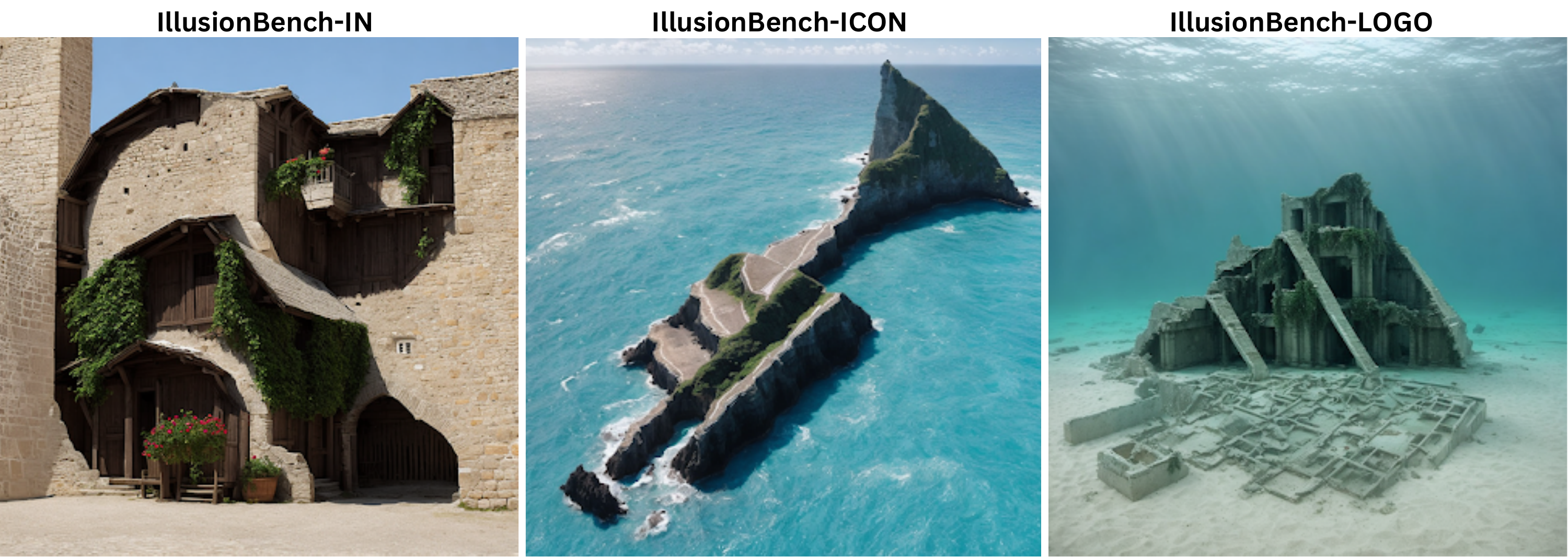}
    \caption{\textbf{Can vision-language models recognize these shapes?} 
    % We evaluate if various vision language models can recognize abstract shapes in \datasetname. 
    \datasetname dataset contains images in which scene elements are arranged to represent abstract shapes. 
    % We present three datasets each with a different embedded shape category. 
    % We observe that even state-of-the-art vision-language models fail to recognize the shapes.
    % Please refer to the original images in the dataset with the original resolution for evaluation.
    }
    \label{fig:demo}
\end{figure}
To address these limitations, we introduce \datasetname,\footnote{
    We use ``Illusion'' in the name of our benchmark because images in our dataset can be understood as instances of pareidolia, an illusion caused by the tendency of the human visual system to identify familiar shapes in complex scenes. Our dataset should not be confused with HallusionBench \cite{guan2023hallusionbench}, which instead serves as a diagnostic tool to distinguish between VLM reasoning error modes, such as those caused by the language component versus visual component of VLMs.
} which represents shape information by an arrangement of visual elements existing in coherent, naturalistic scenes (see \cref{fig:demo}).
% We leverage enerative model ControlNet \cite{zhang2023addingControlNet} with Stable Diffusion \cite{rombach2022highSD}.
% (i.e. a binary image as condition input). 
% {\red ... 1-2 sentence description of (a) how we generated the dataset and (b) summarizes how/why it resolves the issues of the prior datasets ...}
We evaluate vision-language models (VLMs) using \datasetname{} in three scenarios: (1) measuring \textbf{zero-shot} performance of 
% contrastive VLMs (e.g., CLIP ~\cite{radford2021learning}) and 
generative VLMs (e.g., LLava ~\cite{liu2024llavanext}, GPT-4o \cite{openai2023gpt4}, and Gemini ~\cite{team2023gemini}); (2) measuring \textbf{few-shot} performance of VLMs using in-context learning (e.g., \citep{zhao2023mmicl}); and (3) \textbf{fine-tuning} contrastive VLMs (e.g., CLIP ~\cite{radford2021learning}) to recognize abstract shapes and testing their ability to generalize to unseen scenes.
% using samples representing the same abstract object whose presence is created across multiple scenes allows the models to learn to identify such objects in previously unseen scenes. 
% {\red TODO: explain how this resolves the limitations of prior work discussed above}
% \textcolor{cyan}{add all models used in intro}
%
% {\org We introduce three benchmark datasets evaluating VLMs' abstract shape recognition abilities in the settings described above.}
% Our datasets allows for the quantitative measurement of VLMs' shape perception {\olv robustness under superficial visual variations}.
% Differently from existing benchmarks, \datasetname is designed around the natural-language description and question-answering abilities of modern VLMs, querying them for both the type of scene and abstract shape portrayed in generated images.
We find that, while human annotators can easily identify these shapes, VLMs struggle to identify shapes and instead focus on the scene components, failing to exhibit the abstract shape recognition capabilities that are essential for enabling humanlike visual robustness.

\section{Background and Related Work}
\label{sec:rel_works}

% {\red suggestion: let's have this section be titled ``background and related work'' or ``preliminaries'' or something, given that we're also covering human visual perception and pareidolia, neither of which are really ``related work''}

\paragraph{Shape perception and visual recognition}\label{sec:relwk_shape}

Shape information is widely considered to be the most important cue leveraged by the human visual system for object recognition \cite{landau1988importance,biederman1988surface,xu2004emergence,elder2009cue,baker2018abstract}. % object recognition, learning, and perception... "recognition" is most normal in context of CV; but most of the psych lit focuses on (perceptual) learning
Our ability to perceive shapes is crucial in enabling the robustness of human visual perception \cite{hummel2001complementary,ommer2013role},
as shape is invariant to key transformations such as translation, rotation, scaling, and changes in illumination, color, and texture \cite{ommer2013role,kendall1984shape,dryden2016statistical}.
%
% {\red TODO (Adam): 
% % \- explain the importance of shape recognition in human vision, and its importance in facilitating robustness/invariance
% \- briefly touch on gestalts and perceptual grouping (and link to pareidolia -- merge with pareidolia paragraph below?)} 
% \textcolor{cyan}{ash: yes would be great to merge with pareidolia}
%
Thus, many works have investigated the extent to which neural object classifiers rely on shape for visual recognition tasks, finding that early supervised deep neural networks rely more on texture cues rather than shape \citep{geirhos2018imagenettrained,islam2021shape, Benarous2023HarnessingSD, Pinto2022AnIT,subramanian2023spatialfrequency}.
More recently, \citet{gavrikov2024vision} showed that multimodal vision-language models can be prompted to rely more on shape in visual recognition. 
% {\olv
% The images in these experiments contain a shape outline (e.g. a cat) and a texture of a different entity (e.g. an elephant) is applied to the entire image using neural style transfer. 
% However, our proposition evaluates shape recognition from a different perspective. We represent the shape information with a complex arrangement of elements in a complicated scene comprised of various textures and objects. Recognizing the shape in \datasetname requires \emph{holistic} understanding of the image information.
% }
% 
% \paragraph{Shape recognition benchmarks}
% {\org
Each of these works evaluates shape perception on the basis of the Cue Conflict (CC) or Stylized-ImageNet (SIN) benchmarks \cite{geirhos2018imagenettrained};
% in which the texture of an image is changed using style transfer while holding their shape constant.
% This allows one to analyze classifiers' reliance on shape versus texture by testing how often they either (a) classify CC images according to the source shape image versus the target texture image, or (b) change their predictions between original shape images and SIN texture-modified images.
% allowing for downstream analysis of how predictions change in response to changes in texture.
% Humans overwhelmingly % (in >95\% of cases) 
% classify these images according to shape, not texture; but {\red ... merge the preceding text with related work section, "early supervised classifiers relied on texture cues...", and put the following sentence toward the end of the paragraph:}
% While recent work has found that VLMs can be prompted to much more frequently (in 49-72\% of cases, depending on the model and the prompt) classify CC images on the basis of shape rather than texture \cite{gavrikov2024vision}, we find that such strong shape-bias results do not generalize to the datasets we consider.
%
but despite their longstanding utility, we observe several key limitations with these benchmarks:
% in studying shape perception in neural vision models, there are a number of key limitations associated with these datasets that must be addressed:
% Both datasets (CC and SIN; \citealp{geirhos2018imagenettrained}) are generated using legacy style transfer, they exhibit several important limitations {\red ... that can now be addressed with blah ...}:
\begin{compactenum}
    % \item Images generally do not depict coherent visual scenes -- instead, most of the [scene information] other than the image class is lost.
    % \item Textures are applied uniformly to entire images, meaning that the contrast in textures between various objects and the background of any given image is usually lost, yielding perceptually uniform images \cite{chen2021diverse} that lack much of the distinguishing shape and scene information in original images \cite{wang2023stylediffusion}.
    \item 
    % \textbf{Generated images usually do not depict coherent or naturalistic visual scenes.} 
    \textbf{Lack of coherent, naturalistic, and complex visual scenes:} 
    Images contain only the shape of a single class mixed with a single texture applied uniformly to the entire image.
    % \item Textures can be seriously distorted in style transfer, yielding images that appear to neural networks to have the target texture while this texture may not be apparent to humans [see examples from CC; can we find any citations on this or does it have to be ad-hoc qualitative?]
    \item \textbf{Missing shape information:} Key shape information is often lost, yielding ``a substantial fraction'' of images that are unrecognizable to human annotators \cite{geirhos2018imagenettrained}. The contrast in textures between the object and the background of any given image is usually lost, yielding perceptually uniform images \cite{chen2021diverse,wang2023stylediffusion}.
    % that lack much of the distinguishing shape and scene information in original images \cite{wang2023stylediffusion}.
    % meaning that results obtained by testing on these images do not form a meaningful basis for comparison with human vision.
    \item \textbf{Low-quality style transfer:} The style transfer methods in these datasets \cite{gatys2016image,huang2017arbitrary} are known to confuse shape and texture information \cite{wang2023stylediffusion} 
    and often fail to capture fine-grained textures \cite{wang2021evaluate}.
    % \item \textbf{Simplistic images.} 
    
    % {\olv
    % The images in these experiments contain a shape outline (e.g. a cat) and a texture of a different entity (e.g. an elephant) is applied to the entire image using neural style transfer. 
    % However, our proposition evaluates shape recognition from a different perspective. We represent the shape information with a complex arrangement of elements in a complicated scene comprised of various textures and objects. Recognizing the shape in \datasetname requires \emph{holistic} understanding of the image information.
    % }
\end{compactenum}
% As such, we believe it is necessary to create new benchmarks for this task better suited to a new generation of vision models (in part, by using a new generation of generative models), 
To address these limitations, we introduce \datasetname, which leverages state-of-the-art generative models to create images representing shape information with a complex arrangement of elements in detailed visual scenes comprised of various textures and objects. 

\paragraph{Evaluating visual capabilities of VLMs}
Vision-language models (VLMs) have exceeded conventional benchmarks, often even exhibiting capabilities that they are not explicitly trained for ~\cite{bubeck2023sparks} and underscoring the need for new forms of evaluation ~\cite{zhang2024mmllm}. Traditional image recognition benchmarks are not designed to characterize such capabilities, indicating the need for innovative evaluations.
For instance, 
~\citet{bitton2023breakingwhoops} studies commonsense visual reasoning by testing whether models perceive peculiar content in visual scenes; ~\citet{fu2024mme} evaluates VLMs on recognizing the count of objects, relative positions of objects, OCR, and commonsense visual reasoning; and ~\citet{tong2024eyes} 
% points to several visual limits of vision-language models by 
proposes visual tasks requiring fine-grained understanding of object orientation, perspective, and the states of objects in the image.
% ~\citet{bitton2023breakingwhoops} take commonsense visual reasoning to the next level by checking if models perceive the peculiar content in the images. 
Finally, ~\citet{zhou2023analyzingHallucination,li2023evaluatingPope} focus on limitations specific to generative VLMs, such as visual hallucination.

% \subsection{Related Benchmarks} 
% \begin{itemize}
%     \item https://arxiv.org/pdf/2401.11943
%     \item https://arxiv.org/pdf/2212.08044
% \end{itemize}

\section{Benchmark Description}\label{sec:benchmark}
%\textcolor{red}{Todo: put references to appendix and insert examples, e.g. .??? } \fp{What the heck is this?}
\subsection{Generative Process and Notation}

\begin{figure}
    \centering
    \includegraphics[width=\textwidth]{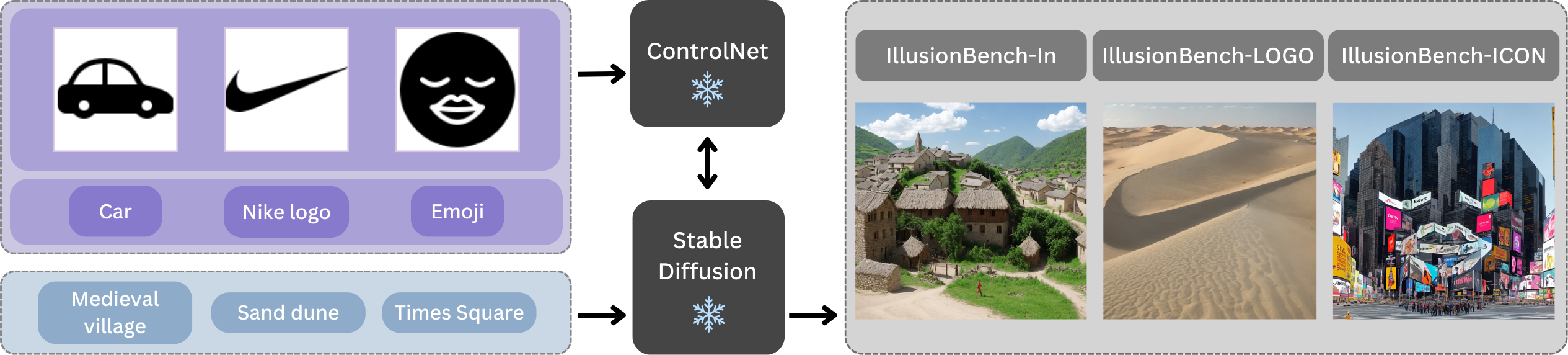}
    \caption{\textbf{Dataset generation.} For each of the 3 datasets in \datasetname, we show an example image from the dataset alongside an example scene prompt and an example shape conditioning image used to generate it. A shape image $x_{i}$ (with the class name $c_{i}$) and a scene description $s_{j}$ are combined to generate the \datasetname image $x_{ij}$.}
    \label{fig:dataset}
\end{figure}

Consider the set $\mathcal{C} = \left\{\left(x_{i},c_{i}\right)\right\}_{i=1}^{|\mathcal{C}|}$ of binary shape conditioning images $x_{i}$ 
representing the shapes of corresponding object class $c_{i}$,
and $\mathcal{T} = \left\{\left(s_{j}\right)\right\}_{j=1}^{|\mathcal{T}|}$ is the set of prompts where each $s_{j}$  
describes a different scene
(e.g., \texttt{Ocean} or \texttt{Medieval Village}).
To synthesize our dataset, we use ControlNet ~\cite{zhang2023addingControlNet}, a module that is trained to control the generative process of text-to-image diffusion models (such as Stable Diffusion; ~\citealp{rombach2022highSD}) by 
conditioning on inputs specifying spatial information to guide the generative process,
such as our shape conditioning images $x_{i}$ (refer to \Cref{fig:dataset} for an overview).
The pipeline (\Cref{fig:dataset}) transforms the tuple $\left(x_{i},s_{j}\right)$ into an image $\x{ij}$ representing the considered shape $x_{i}$ of class $c_{i}$ embedded in a scene of type $s_{j}$.\footnote{
    The generation is conditioned on additional hyperparameters that allow us to obtain shapes that can be recognized at varying levels of abstraction. See \cref{sec:app_hyperparams} for further details.
} We therefore obtain our datasets by creating a tuple $\left(\x{ij},c_{i}, s_{j}\right)$ for each combination of conditioning images and prompts.
We then consider three predictive tasks a VLM $f$ should perform (where $p_C, p_S,$ and $p_{C,S}$ represent prompts querying for $c_{i}, s_{j}$, or both, respectively):
\begin{compactenum}
    \item $\tc$: predict the shape $c_{i} = f(\x{ij},p_C)$.
    \item $\ts$: predict the scene $s_{j} = f(\x{ij},p_S)$.
    \item $\tb$, predicting both the shape and the scene $\left(c_{i}, s_{j}\right) = f(\x{ij},p_{C,S})$.
\end{compactenum}

\subsection{Dataset Details}
\label{sec:data_details}
As exemplified in \Cref{fig:dataset}, the \datasetname benchmark contains three different constituent datasets: \sindata, \logos, and \icons. The number of samples, classes, conditioning images, and domains for each dataset are provided in 
\cref{tab:datasets} 
(with more detailed metadata available in \Cref{app:extra_data_details}).

\begin{table}[h!]
\caption{
% \textbf{\datasetname Splits}. 
Size of each dataset in \datasetname.}
\centering
\begin{tabular}{>{\raggedright\arraybackslash}p{4cm}cccc}
\toprule
Dataset Name     & \# Samples & \# Classes & \# Conditioning Images & \# Scenes \\ \midrule
\sindata   & 6864       & 16         & 48                     & 11         \\
\logos       & 5577       & 21         & 39                     & 11         \\
\icons       & 20064      & 6          & 456                    & 11         \\ 
% \hdashline
% Human Annotation       & 1936      & 32          & 44                    & 11         \\ 
\bottomrule
\end{tabular}
% \vspace{0.3cm}
\label{tab:datasets}
\end{table}

\paragraph{\sindata} 

% Inspired by \citep{geirhos2018imagenet}, we select 16 classes drawn from the WordNet \citep{wordnet} tree employed by ImageNet \citep{imagenet}. {\red this is basically the whole wordnet tree, no? I think we need to be more specific, if possible; otherwise just say "we picked 16 imagenet classes, and for 4 of them where [INSERT REASON], we substituted for near co-hyponyms" (outlined below)
% } 

% We begin with the 16 classes from the most widely-utilized shape perception benchmark, Stylized-ImageNet (SIN) \cite{geirhos2018imagenettrained}.
% Since our benchmark focuses on the detection of shapes emerging from a scene, in order to avoid ambiguities that might make the model hallucinate shapes, we need a set of classes that are characterized by shapes that are visually distinct and easily recognizable {\red justify why}. 
% % and unlikely to naturally occur by random chance (e.g., a closed book is a shape that is likely to occur at random in scenes, since it is a box; on the other hand, a dog silhouette is unlikely to naturally occur). 
% For this reason, we substitute 4/16 of the classes used in \citep{geirhos2018imagenet} that do not meet these criteria for near co-hyponyms, and collect 3 shape conditioning images for each class (see \cref{app:extra_data_details} for further details). 
% % Finally, for each such class, we collect 3 shape conditioning images.

We build upon the 16 classes from the most popular shape perception benchmark, Stylized-ImageNet (SIN) \cite{geirhos2018imagenettrained}. However, since we are interested in how well models can find shapes within a scene, we need clear and distinct shapes that can be identified unambiguously.
% Indeed, ambiguous shapes could trick the model into "hallucinating" the presence of objects a human would not detect. % that aren't actually present.
To address this, we replace 4 of the 16 SIN classes with similar categories (near co-hyponyms) with more distinct shapes. We collect 3 conditioning images for each class.
% (see \cref{app:extra_data_details} for further details). 

\paragraph{\logos} 
% While the \sindata split consists of simplified representations of real-world objects, these simplified shapes occur very infrequently in the real world and mostly as part of infographics or iconic communication. On the other hand, there is a natural category of symbols that are deliberately designed to have a recogniseable and simple shape and frequently occur in real scenes: logos. For this purpose, we collect X different logos. 
% \fp{I think this motivation is very weak}
Another category of shapes that are specifically designed to be visually distinct and easily recognizable are logos,
% To expand easily-integrated domain, we introduce an 
which provide an interesting contrast to the shapes in \sindata, as recognizing them requires 
world knowledge specific to the category of product brands (rather than culturally-nonspecific real-world object classes).\footnote{
    Given that this task requires both world knowledge of product brands \emph{and} abstract shape recognition capabilities, and considering that our goal with \datasetname is only to evaluate the latter, we normalize scores by averaging results for each VLM exclusively on samples obtained from raw shapes that the VLM can recognise in a zero-shot setting, meaning that models are not penalized for lacking world knowledge of specific brands. See \cref{app:individual_dataset_results_zeroshot,app:individual_dataset_results} for non-normalized results by class.
}
% knowledge of cultural information like product brands rather than general semantic categories like ImageNet/WordNet classes.
% rather than a collection of generic synsets from WordNet.
Thus, we expand our dataset to this domain by collecting 39 different logo conditioning images across 21 brands.
% (see \cref{app:extra_data_details} for further details).

% \paragraph{\icons (Coarse Icons)} 
\paragraph{\icons} 
Finally, we develop a third dataset to test whether VLMs can be trained to recognize cross-modal abstractions over perceptually distinct shapes representing semantically related concepts (e.g., where images representing shapes of owls or turtles are both recognized as instances of the ``animal'' class, despite having very different shapes).
We create a coarse-grained dataset of 6 (informal) hypernym categories across 456 emojis as shape conditioning images.

% {\red TLDR: we want more than one conditioning image per class -- more variety, complexity; coarser form of classification, more in-class diversity. we do this for emojis because it's super easy to collect them based on emoji category (e.g., faces, animals, etc.)}

%\paragraph{}
%{\red TODO: briefly add annotation info (how many annotators, how many annotations, randomly sampled from each, accuracy range) + refer to appendix for more details}
\paragraph{Validating Dataset Quality}
Although ground truth labels for object classes and scene types are available, image generators may sometimes produce low-quality or high-difficulty images whose object shape is not human-recognizable.
To minimize the proportion of such images, we begin by restricting the hyperparameters that control the influence of the conditioning image to ranges that we qualitatively found to produce clearly distinguishable shapes (see \cref{sec:app_hyperparams}). 
To validate that the shapes in the resulting images are indeed human-recognizable, we recruited 60 participants (information is anonymized) to manually annotate randomly sampled subsets of \sindata, \logos, \icons, obtaining an average annotator accuracy of 95.6\%, 97.17\% and 96.8\%, respectively, indicating that humans are indeed able to recognize the shapes in the vast majority of the generated images.\footnote{
    Note that human annotator accuracies are only intended to \emph{validate the quality of the generated dataset} and \emph{confirm that the resulting abstract shapes are indeed human-perceptible}. They are \emph{not} intended for direct comparison with VLM performance, as there are a few fundamental differences in 
    how annotators and VLMs are tested. For instance, where VLMs do not know the purpose or structure of the task beyond what is included in the prompt, annotators are shown onboarding materials describing the task, including several pre-annotated examples.
    % including 5 pre-annotated examples to teach them the task, and further receive feedback on their performance for 10 additional test images (in order to ensure that they understand the task). As such, annotator performance is more similar to few-shot evaluation rather than zero-shot, with the additional caveat that annotators are provided with much more information and context about the task they are performing than could be included in prompts (e.g., how the dataset is constructed and why, individualized feedback on the initial 10 test images, etc.), effectively ``priming'' them to recognize shapes where they might otherwise have perceived only the background scene.
} (See \cref{app:annotation} for further details.)
%this appendix and we're happy if we survive the day :) {\red how many from each? what is statistical significance? kendall's tau?}

\subsection{Evaluation}
\label{sec:evaluation}
%{\red when writing this section, treat ICL and zero-shot as having same eval (even though slightly different at present), and link to appendix where we do actually spell out the difference}

%{\red
%TODO  (Adam): 
% \- write up the procedure for constrained decoding
% \- justify CD against ``response extraction'' methods from \cite{gavrikov2024vision}) -- TLDR (needs rephrasing): As they mentioned, this is BERT- or CLIP-style analysis that was built around the era before generative models where all you could do to compare an open set of class options was to compute cosine similarities between embeddings of query (e.g., image or question) and possible answers. Doing this with generative models like the VLMs we consider is not reasonable, since we can just directly compare probabilities of generating different answers (and also, generative models aren’t even trained for this, unlike contrastive models like CLIP).
%\- (probably in appendix) make a point about the difference between \citet{gavrikov2024vision}'s metrics and ours -- basically, we add failure cases to the denominator of (our analog of) shape and texture bias
%}

% \item $\tc$: predict the shape $\yc{i} = f(\x{ij},p_C)$.
%     \item $\ts$: predict the scene $\yt{j} = f(\x{ij},p_T)$.
%     \item $\tb$, predicting both the shape and the scene $(\yc{i}, \yt{j}) = f(\x{ij},p_{C,T})$.

Given image $x_{ij}$, we prompt VLM $f$ with both $x_{ij}$ and prompts $p_k$ corresponding to the shape, scene, and both the shape and scene (i.e., where $p_k$ is variously $p_C, p_S,$ or $p_{C,S}$, respectively), yielding responses $r_k = f(x_{ij}, p_k)$ for each prompt $p_k$.
% check if anywhere in the response of the model either $\yc{i}$ or $\yt{j}$ is present. 
For each $x_{ij}$, we evaluate \emph{shape recall} on the basis of whether the term $c_{i}$ appears in the response $r_C$ or $r_{C,S}$ (yielding $1$ if so, or $0$ if not), and evaluate \emph{scene recall} by whether $s_{j}$ appears in $r_T$ or $r_{C,S}$ (similarly yielding $1$ or $0$), and report the shape and scene recall for each dataset as the sum of the recall figures across all $x_{ij}$ instances divided by the size of each dataset.
% consider the answer to be correctly predicting the shape; and if $\yt{j}$ is present, we consider the answer to be correctly predicting the scene. 
% We average shape and scene recall over each dataset, therefore introducing two metrics: \emph{shape accuracy} and \emph{scene accuracy}, respectively. 
In contrast to prior related works (e.g., \citealt{geirhos2018imagenettrained,geirhos2021partial,gavrikov2024vision}), our proposed metrics are designed such that shape recognition performance is not in competition with the ability to recognise other visual elements (e.g., textures or scene elements),
% Indeed, differently from closed-set classifiers considered by \cite{geirhos2018imagenettrained}, which must select only one among a pre-defined set of options, modern VLMs can generate detailed descriptions of images, capturing nuanced information about shape and scene at the same time.  For further discussion, refer to \cref{app:metrics}. 
as -- unlike traditional classifiers, which must select only one among a pre-defined set of discrete classes -- generative VLMs can respond with detailed descriptions of images including information about shape, scene, or other visual elements at the same time (or given different prompts).

\subsection{Experimental Overview} \label{sec:use-cases} In the following sections, we evaluate the shape perception capabilities of modern VLMs on \datasetname under the following paradigms: 
% The section delves into a series of experiments designed to answer three key research questions:
\begin{compactitem}
    \item \textbf{Zero-Shot Recognition}: Given that an instruction-tuned VLM can recognise a shape $x_{i}$, can it identify the same shape when it emerges from the combination of visual elements in $x_{ij}$ without any explicit examples or specialized fine-tuning? (\cref{sec:zs_exp})
    \item \textbf{Few-Shot Learning}: Given that a multi-modal in-context learner can recognise a shape $x_{i}$ zero-shot, can it leverage few examples to learn to identify it in $x_{ij}$? (\cref{sec:icl_exp})
    \item \textbf{Domain Generalization}: Given training samples $\{ x_{ij} \}$ representing a shape $x_{i}$ in certain types of scenes, can models learn to recognise the same shape in other, unseen scene types? (\cref{sec:mdg_exp}) %Can models learn to recognise shapes given some training samples characterised by specific scene prompts,  representations of shapes across  training domains characterised by   $\mathcal{D}_j$ to learn representations invariances across different representations $\xc{ij}$ of the same shape $\xc{i}$  to learn generalizable features that function even in previously unseen domains? 
    % Can we use Domain Generazation instead of Multi-Domain Generalization?
\end{compactitem}
% Through these experiments, we aim to shed light on the strengths and limitations of current VLMs in comprehending and recognizing complex visual stimuli that emerge from the interplay of multiple elements within a scene.

% \section{Evaluation Metrics}
% \input{Arxiv/chapters/metrics}

\section{Can Instruction-Tuned VLMs Recognize Shapes Zero-Shot?}
\label{sec:zs_exp}

% \begin{wrapfigure}{r}{0.6\textwidth}
%     \centering
%     \includegraphics[width=0.58\textwidth]{results/Total_plot_shape_v12.pdf}
%     \caption{\textbf{Performance Comparison of Various VLMs}. This plot illustrates the performance of VLMs in terms of shape and scene recognition accuracy. Based on this plot, we observe a significant bias towards scene recognition in our datasets. Also, the performance of the VLMs in previous work (Stylized ImageNet) demonstrates a notably better capability in abstract shape recognition.}
%     \label{fig:shape-zero-shot}
% \end{wrapfigure}

\begin{figure}
    \centering
    \includegraphics[width=\textwidth]{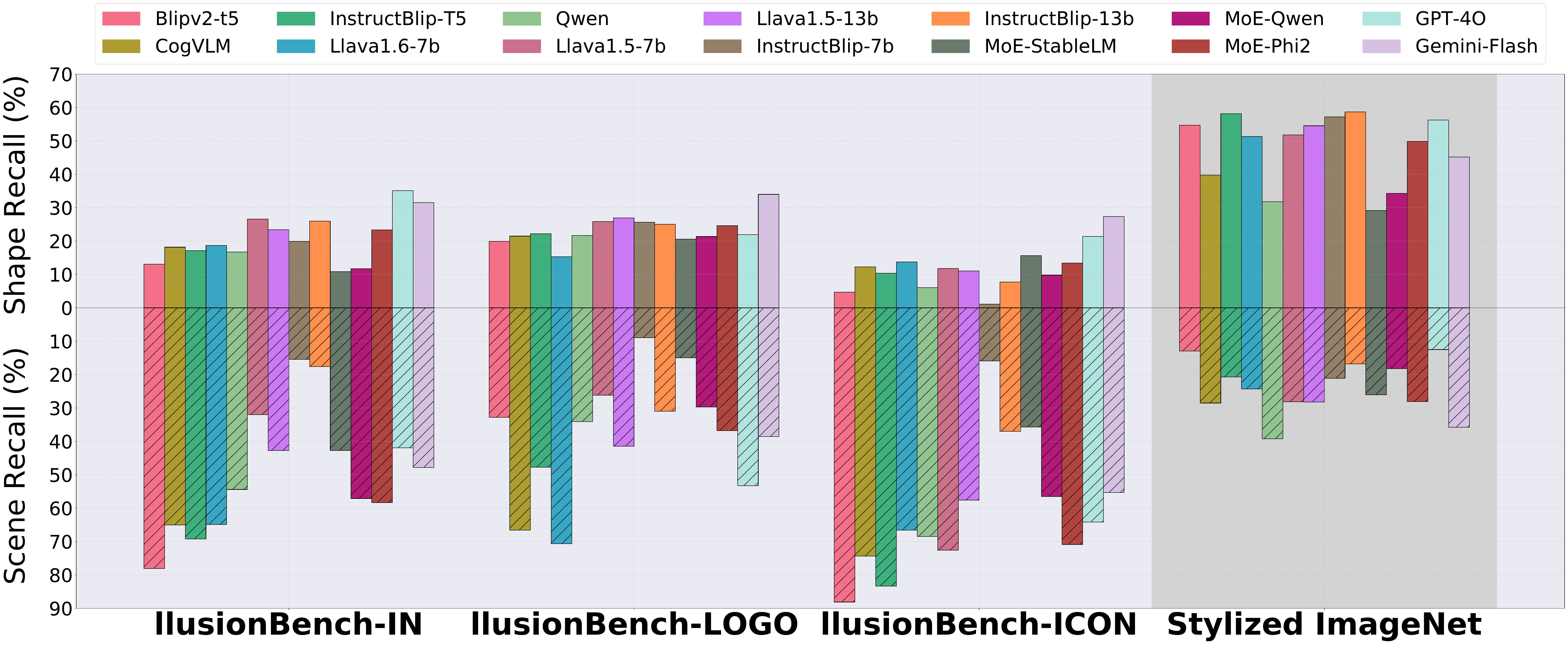}
    \caption{\textbf{Zero-Shot Results}. Average shape and scene recall of VLMs across each \datasetname dataset, compared with Stylized-ImageNet \cite{geirhos2018imagenettrained} (rightmost, shaded).}
    \label{fig:shape-zero-shot}
\end{figure}

% {\red ~\\ TODO:
% \- resize text in figure so it ends up being similar size as paper text on the page
% \- standard y axis in both directions -- let's go up to 100\% in both directions
% \- question: SIN doesn't have a "scene acc", or even a "texture bias", per se -- is this supposed to be cue conflict? also, why is it so low
% }

%Considering the impressive growth of Vision-Language Models (VLMs) and the results obtained in related works ~\cite{gavrikov2024vision} on testing previous datasets ~\cite{geirhos2018imagenettrained}, it is evident that Abstract Shape Recognition in VLMs is significantly more powerful compared to Vision Transformers (ViT), Contrastive Language-Image Pre-training (CLIP), and Convolutional Neural Networks (CNN). Additionally, results from previous dataset tests and compared with the VLMs results on the previous dataset confirm that cutting-edge models have demonstrated substantial improvements in Abstract Shape Recognition. In our evaluations, and to illustrate the weaknesses of VLMs, we have conducted tests on these models. This section elaborates on these tests and the models evaluated.

% 

\paragraph{Experimental Design} 
% The models are prompted to identify the object shape, logo or icon type that is integrated in a scene whose elements contribute to forming the image.
In this experiment, we prompt VLMs zero-shot to identify the abstract shape represented in a visual scene among a closed set of object classes.
We begin by testing whether models can correctly classify the shape conditioning images (binary shape images), and generate images for \datasetname exclusively using these condition shapes.
We then prompt models with respect to the shape and scene in each generated image, and measure the corresponding recall metrics as described in \cref{sec:evaluation}.
(See \cref{app:fullzeroshot} for additional details regarding the experimental design, prompts, and models used in this experiment.)

\paragraph{Models}
% {\red TODO: list and cite models}
We consider the following VLMs for evaluation:
\texttt{GPT-4o} \citep{openai2023gpt4}, \texttt{Gemini-Flash} \citep{team2023gemini}, \texttt{LLaVA1.5/6-7/13b} \citep{liu2024visual}, \texttt{CogVLM} \citep{wang2024cogvlm}, \texttt{BLIPv2-t5} \citep{li2023blip2}, \texttt{InstructBLIP-7/13b} \citep{dai2024instructblip}, \texttt{Qwen-VL-Chat} \citep{bai2023qwen}, and \texttt{MoE-StableLM/Qwen/Phi2} \citep{lin2024moellava}. 

\paragraph{Results } 
Our main findings in this experiment (visualized in \cref{fig:shape-zero-shot}) are as follows:
\begin{compactitem}
    \item
    For each of our datasets, shape recall is quite low, with most models ranging between 10-30\% (in contrast to the previous dataset, Stylized-ImageNet \cite{geirhos2018imagenettrained}, where all fourteen models exceed 30\%). 
    \item 
    % \textit{Models generally exhibit better scene recognition than shape recognition abilities.} 
    For nearly all models and datasets, models exhibit superior scene recall relative to shape recall. % with few exceptions such as \texttt{InstructBLIP-7/13b}
    This indicates that the recognition capacity of current VLMs is still biased towards scene/texture features, similar to earlier work studying CNN classifiers (see \cref{sec:relwk_shape}).
    % {\red list 1-2 implications}
    % With few exceptions, models generally exhibit superior scene recall relative to shape recall.
    % {\red TODO: in a separate bullet, list the exceptions, and let's briefly analyze them}
    %\item Models simultaneously show worse shape recall and better scene recall on \icons than \sindata or \logos. {\red any hypotheses as to why? }
    \item \texttt{GPT-4o} and \texttt{GEMINI} show superior shape recall to all other models in 3/3 and 2/3 of our datasets, respectively, demonstrating a shape-recognition gap between the best available open- and closed-source VLMs.
    % in terms of having high shape accuracy whilst simultaneously having low scene accuracy on both ICONS and LOGOS. This shows that \texttt{GEMINI} generally follows/understands what it needs to predict (shape) better than the other models in our experiments.
    % \item \texttt{GPT} shows a more dramatic shift in preference to scene prediction than \texttt{GEMINI} moving from Stylzed-ImageNet to our dataset.
    \item Mixture of Experts (MoE) (like \texttt{MoE-StableLM}, \texttt{MoE-Qwen}, and \texttt{MoE-Phi2}), which are generally employed to improve models' performance, exhibit neither superior shape nor scene recall with respect to individual models or closed-source models. 
    \item Among all open source models, \texttt{LLava} attains the strongest shape recall performance across all our datasets. In contrast, \texttt{Blipv2} attains the highest scene recall (except for the \logos split).
\end{compactitem}

See \cref{app:fullzeroshot} for more fine-grained results and analysis.

\section{Can In-Context Learners Learn to Identify Abstract Shapes?}
\label{sec:icl_exp}

Given zero-shot prompting exhibits poor performance at detecting abstract shapes and shows VLMs mostly focus on background stimuli, a natural question is whether it is possible to teach models to recognise known shapes with a few samples by leveraging their In-Context Learning (ICL) or few-shot capabilities.\footnote{See \Cref{app:icl-description} for a brief introduction to ICL.} 

\paragraph{Experimental Design}  We restrict our experiments to samples generated from conditioning images $x_{i}$ that models can correctly classify in a zero-shot fashion (see \cref{app:icl-experimental-details}). %{\red justify this or link to appendix about it} %(see \Cref{sec:use-cases} \tom{check correct reference} for further discussion). 
Let us focus on the predictive task $\tc$ (as analogous formulations of ICL apply for $\ts$ and $\tb$). Given that the model can correctly assign the class $c_{i}$ to the conditioning image $x_{i}$, we provide it with the context sequence $\left\{(\x{{i_{w}},{j_{w}}}, c_{i_{w}})\right\}_{w=1}^{|W|}$, where $W$ is the context window plus a test image $\x{{i_{*}},{j_{*}}}$, and prompt the model to predict the object's shape $c_{i_{*}}$.  

Using specific constraints on context sampling relative to a test sample, we define four learning tasks corresponding to perceptual challenges:

\begin{compactitem}
\item \textbf{ICL1}: \emph{Given the context lacks any image depicting the scene or shape type of the test sample $\x{{i_{}},{j_{}}}$, can the model recognize its shape $c_{i}$?}
\item \textbf{ICL2}: \emph{Given the context includes an image of the shape type but not the scene type of the test sample $\x{{i_{}},{j_{}}}$, can the model recognize its shape $c_{i}$?}
\item \textbf{ICL3}: \emph{Given the context includes an image of the scene type but not the shape type of the test sample $\x{{i_{}},{j_{}}}$, can the model recognize its shape $c_{i}$?}
\item \textbf{ICL4}: \emph{Given the context includes images of the scene type and shape type of the test sample $\x{{i_{}},{j_{}}}$ (separately and exactly once), can the model recognize the test sample's shape $c_{i}$?}
\end{compactitem}

\begin{figure}[t]
    \centering
    \includegraphics[width=\textwidth]{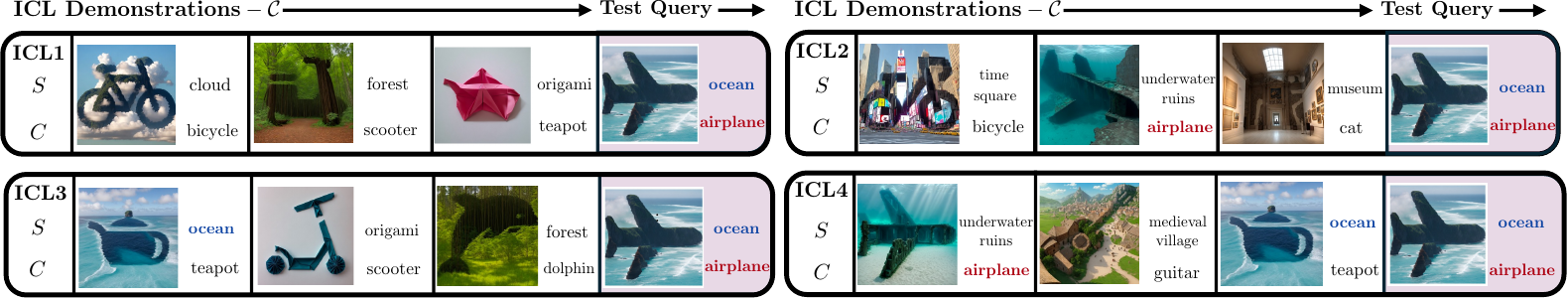}
    \caption{\textbf{ICL Learning Tasks}. Figure depicting the four ICL learning tasks, $ICL1, ICL2, ICL3$ and $ICL4$,  defined by constraints on demonstration example selection as introduced \Cref{sec:icl_exp}.}
    \label{fig:icl-demo-selection}
\end{figure}
Samples in the context are selected uniformly at random, excluding those that do not satisfy the constraints for a given test sample. Random selection serves as a simple baseline for ICL example selection, avoiding confounding factors like similarity bias or majority  \citep{bertini2024makes}. We perform $0, 1, 2, 4, 8$-shot on \logos and \sindata, and $1, 2, 4, 5$-shot on \icons. Further details of ICL experiments can be found in \Cref{app:icl-experimental-details}. We additionally perform ablations to examine the sensitivity of our results to the prompt template used or to the order in which in-context examples are given to the model. These additional results can be found in \Cref{app:ICL ablations}.

\paragraph{Models.} We consider several state-of-the-art models that have been designed to support ICL: (1) \texttt{LLaVA-Next} \citep{liu2024llavanext}, (2) \texttt{Qwen-VL-Chat} \citep{bai2023qwen}, (3) \texttt{Otter-MPT} \citep{li2023mimic}, (4) \texttt{IDEFICS-9B-Instruct} \citep{laurenccon2024obelics}, and (5) \texttt{MMICL-T5-XXL}\citep{zhao2023mmicl}. (We describe each models, the prompts they are provided, and a detailed motivation for selecting these particular models in \Cref{app:icl_models}.)%  {\red can we add a brief explanation of why these, and only these, models? (e.g., why not all the models applied in zero-shot setting -- is it context window, poor ICL pilot performance, ...?)}

\begin{figure}
    \centering
    \includegraphics[width=\textwidth]{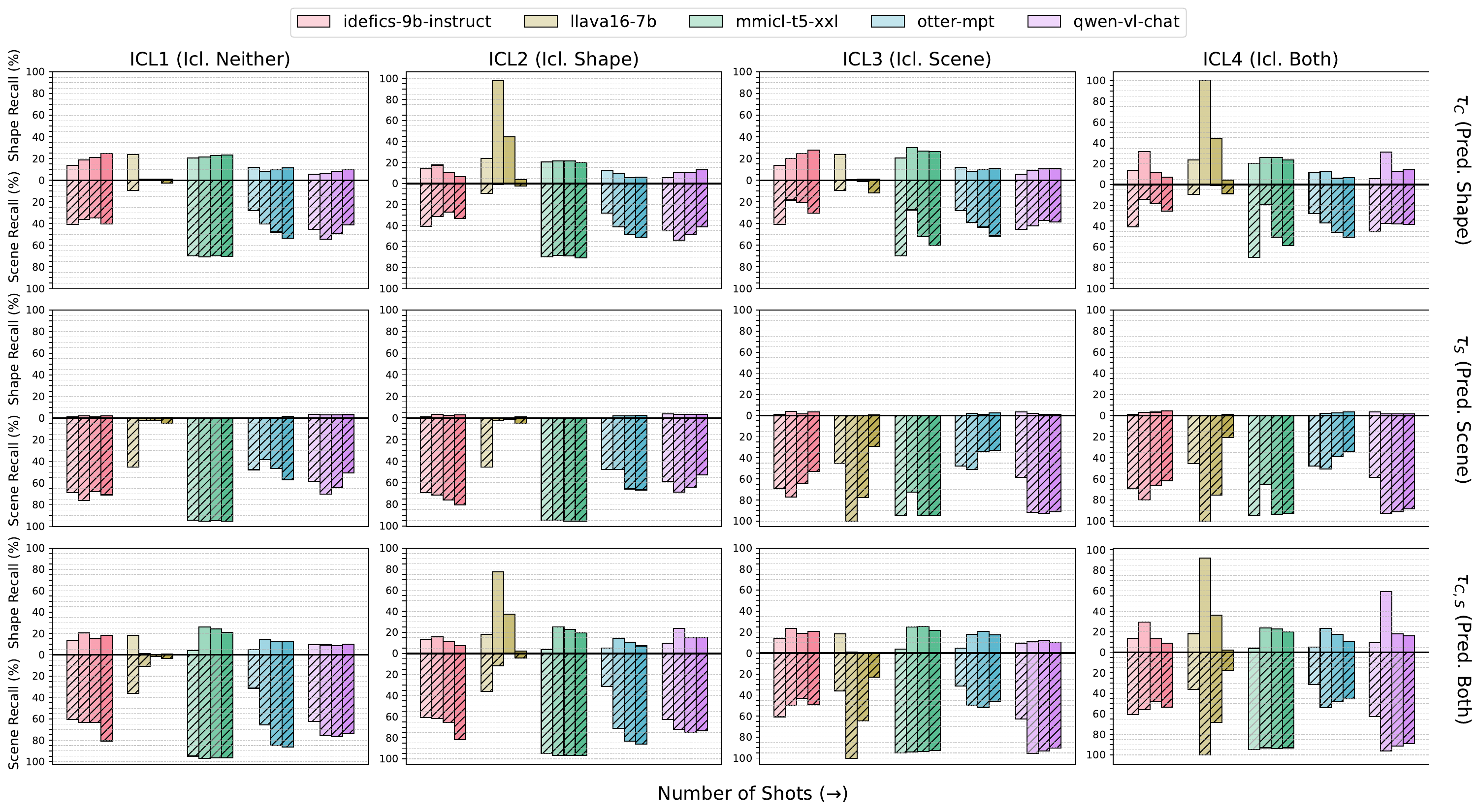}
    \caption{\textbf{ICL Results}. Few-shot (0,1,2 and 4-shot) shape and scene recall of VLMs averaged across the \logos, \sindata and \icons datasets, displayed for the different ICL learning tasks and the different prediction tasks.
    }
    \label{fig:icl_combined}
\end{figure}

\paragraph{Results.} We summarise the average results across all three dataset splits for $0,1,2$ and $4$-shot ICL as  show in \Cref{fig:icl_combined} following the recall metrics introduced in \Cref{sec:evaluation} (for results for individual datasets and for $5$-shot and $8$-shot performance on \icons and \sindata / \logos respectively, see \Cref{app:individual_dataset_results}.) \footnote{In \cref{fig:icl_combined}, we observe that \texttt{LLaVA} shows often close to zero recall on either shape of scene prediction. We explore a few possible reasons for these results in \Cref{app: llava-bad-performance}.} We report here the main trends in the data. Discussion of exceptions that do not follow the reported general trends can be found in \Cref{app:ICL-exceptions}.
\begin{itemize}
\item \textit{ICL does not mitigate tendency to predict scene over shape.} 
As shown in \Cref{fig:icl_combined}, ICL has minimal effect in altering the models' tendency to predict the scene $s_{j}$, regardless of whether the prediction task is $\tc$ (predict shape), $\ts$ (predict scene), or $\tb$ (predict both). 
    \item \textit{On average,} \texttt{MMICL-t5-XXL} \textit{exhibits the strongest scene and shape recall for the highest number of shots}  (i.e., when majority voting biases decay; see \citep{bertini2024makes}). 
  \item \textit{Increasing the number of shots has mixed effects on performance.} We observe in \Cref{fig:icl_combined} that the models often exhibit non-monotonic performance trends for both shape and scene recall across all prediction tasks and demonstration selection constraints. In general, this indicates that the models struggle in general to adapt to tasks $\tc$, $\ts$, and $\tb$, even with increasing demonstration examples. These results are in line with previous findings that complex ICL tasks remain challenging for current visual language models (VLMs) \citep{zong2024vl}. 
    \item  \textit{Context selection strategy effects prediction tasks differently}.
    \begin{itemize}
        \item \textit{$\tc$ (shape prediction):} As shown in the top row of \Cref{fig:icl_combined} for task $\tc$ (shape prediction), including the shape in the context (ICL2 and ICL4) either maintains or reduces performance for most models such as \texttt{MMICL} and \texttt{IDEFICS}. This suggests that most models struggle to identify and disentangle shape from the scene through ICL.
        
        \item \textit{$\ts$ (scene prediction):} The second row of \Cref{fig:icl_combined} shows the mixed effect of including the scene within the context (ICL3 and ICL4) compared to not including it (ICL1 and ICL2). Models such as \texttt{LLAVA}, \texttt{OTTER} show a reduction in scene recall and when including the scene in the context. \texttt{MMICL} maintains comparable performance, whereas $\texttt{LLaVA}$ and $\texttt{QWEN}$ show improved performance. 
        \item \textit{$\tb$ (predicting both shape and scene):} The final row of \Cref{fig:icl_combined} typically shows trends similar to  $\tc$ and $\ts$ -- e.g., scene recall and shape recall for \texttt{MMICL} (whose zero-shot shape recall is lower on this task than in $\tc$), \texttt{IDEFICS}, and \texttt{LLaVA} are comparable with respect to those in $\tb$ and $\ts$ (respectively).
    \end{itemize}    
\end{itemize}

Overall, we observe that ICL does not substantially aid models in learning to detect abstract shapes within scenes or to help reduce scene prediction bias. The non-uniformity of relative results between models further highlights the immaturity of ICL for multi-modal models, particularly for complex tasks like abstract shape recognition. 
% In sum, the limited ICL ability of these models can be understood as a limitation of for our experiments here.

% \section{Can Invariant Representations Across Domains Be Learned?}
\section{Can VLMs Learn Invariant Representations Across Domains?}
\label{sec:mdg_exp}

A compelling application of \sindata lies in Domain Generalisation (DG) \cite{gulrajani2020search}. A visual domain is a set of samples with shared characteristics that influence the appearance of objects (e.g., shared style, such as cartoons, paintings, or photos; shared lighting conditions, such as photos taken at similar times of day with similar weather conditions; etc.). 
In DG, the goal is for models to learn domain-invariant representations -- i.e., generalisable features that are predictive of task labels across any domain -- by training across multiple ``source'' domains and testing how well models generalise to unseen test domains. 
(See \Cref{app:mdg_lit} for a more detailed introduction to DG.)

\begin{figure}[h]
    \centering
    \includegraphics[width=\textwidth]{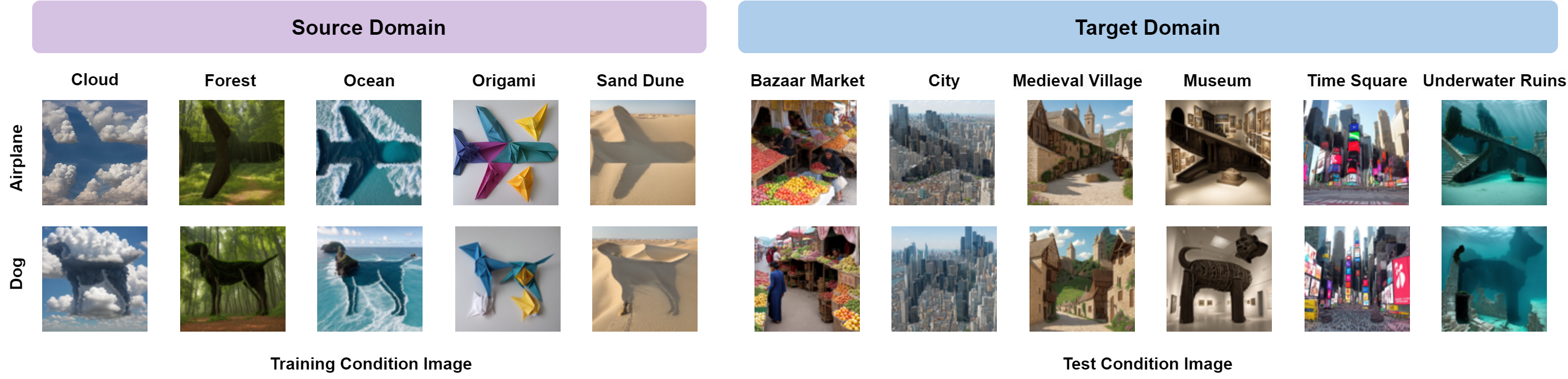}
    \caption{\textbf{\sindata for Domain Generalisation.} We split the dataset into five source domains for training and six target domains for testing. The condition images for generated data samples are only shared among source and target domains, respectively, without overlapping.}
    \label{fig:dg_demo}
\end{figure}

\paragraph{Experimental Design.} We consider all images generated using the same scene prompt $s_{j}$ as coming from the same domain $\mathcal{D}^j$. As shown in \Cref{fig:dg_demo}, we partition the \sindata dataset split into train domains $s_{j} \in \{\texttt{Cloud, Forest, Ocean, Origami, Sand Dune}\}$ and test domains  $s_{j} \in \{\texttt{Bazaar Market, City, Medieval Village, Museum, Times Square, Underwater}\}$. (Conditioning images $x_{i}$ used to generate the training domains are not contained in the test domains.) %This approach prevents the model from depending on the simplicity bias associated with shapes.
We then consider a contrastive language-vision encoder (CLIP \citep{radford2021learning}) and prompt CLIP in order to identify the class $c_{i_{*}}$ of a test sample $x_{i_{*}}$ among all possible shape classes\footnote{Since the zero-shot performance is particularly low, we do not confuse the model further asking it to distinguish the shape from the background type. This also allows us to make the comparison with probing and fine-tuning techniques that deliberately aim at extracting shape more fairly.}. Throughout the experiment, we use ``\texttt{A photo of \{class\_name\}}'' as the prompt template. (See \Cref{app:mdg_hyper} for further experimental details.) 

\textbf{Methods Considered. } We compare various domain generalisation methods including ERM, MixUp \citep{yan2020improve}, RegMixUp \citep{regmixup}, GroupDRO \citep{sagawa2019distributionally}, and VREx \citep{krueger2021out}, using both linear probing and full-parameter finetuning. Besides linear probing, we also consider DPLCLIP \citep{zhang2023domain}, a prompt optimization approach specifically designed for CLIP domain generalisation. 

\begin{figure}[h]
    \centering
    \includegraphics[width=\linewidth]{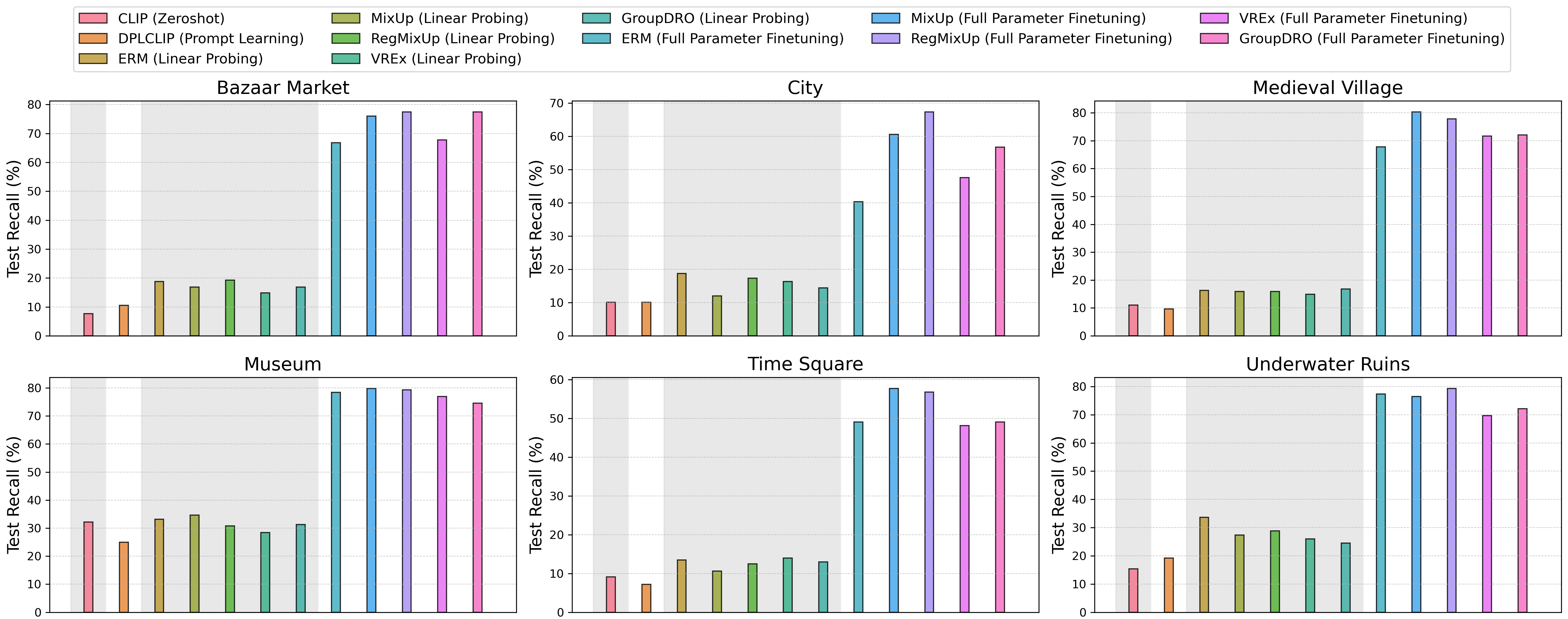}
    \caption{\textbf{Domain Generalisation.} CLIP performance on \sindata for different fine-tuning approaches. Each sub-figure represents an unseen test domain. The categories of approaches, with alternating shading from left to right to indicate these different categories, are: zero-shot prediction, prompt learning, linear probing, and full parameter fine-tuning.} 
    \label{fig:clip_dg}
\end{figure}

\paragraph{Results. } We summarise our findings (reported in \Cref{fig:clip_dg}) as follows:
% \textit{CLIP cannot recognise shapes well in a zero-shot setting.} The CLIP model attains on average extremely low performance in zero-shot settings, with the exception of the \texttt{Museum} domain. This can be attributed to the fact that certain samples within this domain do not simply assemble $c_{i}$ from visual cues of other objects, but sometimes incorporate it as a sculpture. 
% \textit{The CLIP embeddings capture only partially shape information.} Although applying prompt learning for domain generalisation leveraging DPLCLIP is not particularly effective with an average test accuracy of 13.62\%, ERM results particularly effective in improving over the zero-shot performance with accuracy 22.36\% and outperforms all other probing techniques. However, the relatively low absolute values of accuracy indicate the embedding space does not render the test samples linearly separable based on shape criteria. 
% \textit{CLIP can learn features that allow to distinguish objects based on shape using multiple domains.} In full-parameter fine-tuning, it is possible to learn representations that are more oriented towards shape recognition: in all cases a substantial improvement is observed with respect to linear probing. The best performing methods are Mixup and RegMixup, which attain 71.79\% and 73.00\% on average accuracy, respectively.
\begin{compactitem}
    \item \textit{CLIP cannot recognise shapes well in a zero-shot setting.} The CLIP model attains on average extremely low performance in zero-shot settings, with the exception of the \texttt{Museum} domain. This can be attributed to the fact that certain samples within this domain do not simply assemble $c_{i}$ from visual cues of other objects, but incorporate it as a sculpture. 
    \item \textit{CLIP embeddings only partially capture shape information.} %: prompt tuning (or, equivalently, linear probing) can only partly improve the performance.}
    Applying prompt learning for domain generalisation via DPLCLIP is not particularly effective with an average test accuracy of 13.62\%, and ERM results are more effective in improving over the zero-shot performance with accuracy 22.36\%, outperforming all other probing techniques. However, the relatively low absolute values of accuracy indicate the embedding space does not render the test samples linearly separable based on shape criteria. 
    %Applying linear probing methods that leverage the domain generalisation techniques also provide considerable performance boosts. The best linear probing, ERM achieves XXXX average accuracy, RegMixup achieves XXXX and VREX achieves XXXX. Coherently with \cite{DomainBed}, we find that IRM and GroupDRO can underperform with respect to ERM, achieving respectively XXXX and XXX. Nevertheless, the observed performance is far from perfect, therefore indicating that the shape information cannot be easily separated by a linear hyperplane in the representation space of CLIP.  
    \item \textit{CLIP can learn features that allow to distinguish objects based on shape using multiple domains.} In full-parameter fine-tuning, it is possible to learn representations that are more oriented towards shape recognition -- in all cases, a very large improvement is observed with respect to linear probing. The best performing methods are Mixup and RegMixup, which attain 71.79\% and 73.00\% on average accuracy, respectively.
\end{compactitem}

\section{Social Impact}
\label{sec:impact}
The limited shape perception abilities of current vision systems, as highlighted in our work, could hypothetically be exploited by malicious users to, for instance, disseminate hateful or sensitive material online by bypassing inappropriate content visual filters that cannot recognize human-perceptible abstract shapes in scene elements (as enabled by the data-generation methodology we explore in this work).
% whose perceptual modules would not recognize the presence of inappropriate objects camouflaged through complex visual transformations. 
Conversely, improving perception ability could also aid censorship by moderators.
In general, we anticipate that shape recognition capabilities on-par with generative techniques would empower platforms relative to users (e.g., for both content moderation and potential censorship), and shape recognition capabilities that are not able to recognize abstract shapes in outputs of leading generative techniques (as we observe in this work) empowers users relative to platforms, irrespective of whether content is legal or ethical.

\section{Conclusion}
\label{sec:conclusion}
We present \datasetname, a collection of 3 datasets to evaluate shape recognition in vision-language models (VLMs) by representing abstract shapes as complex arrangements of visual scene elements. While human annotators identify these shapes with high accuracy, we find that state-of-the-art VLMs fail to identify the shapes in these scenes zero-shot, tending to focus on scene elements instead. We observe that in-context learning does not significantly improve models' ability to detect abstract shapes;
% in scenes or reduce scene prediction tendency; 
% Using \datasetname we also devise a domain generalization problem setting and show that through fine-tuning, the CLIP model can learn domain-agnostic representations. 
but we do find that contrastive VLMs such as CLIP can be fine-tuned to recognize these shapes and generalize to new scene domains.
% In sum, \datasetname evaluates if the models can see holistically, and the reported limits in models point to directions of improvement for feature research.
%asdf
% We find that, while human annotators can easily identify these shapes, VLMs struggle to identify shapes and instead focus on the scene components, failing to exhibit the abstract shape recognition capabilities of humans that are essential for our visual robustness.
In highlighting the limited shape perception abilities of current VLMs, we hope that \datasetname will help guide future research in developing more robust computer vision systems.
The contributions of each author are listed in \cref{sec:contributions}.

\newpage
\section*{Acknowledgements}
This work is supported in part by the National Science Foundation and the Institute of Education Sciences, U.S. Department of Education, through Award \#2229612 (National AI Institute for Inclusive Intelligent Technologies for Education). Any opinions, findings, and conclusions or recommendations expressed in this material are those of the author(s) and do not necessarily reflect the views of National Science Foundation or the U.S. Department of Education.

Ashkan Khakzar and Philip Torr are supported by UKRI grant: Turing AI Fellowship EP/W002981/1, and by the Royal Academy of Engineering (United Kingdom) under the Research Chair and Senior Research Fellowships scheme.

We would like to express our gratitude to Yawei Li for performing some preliminary experiments on a different topic (not included in this work) before we converged on the research topic explored in this work. We also extend our thanks to Ali Ma'manpush, Julia Hockenmaier, and Prashant Jayannavar for their invaluable assistance and advice regarding the human data annotation process.
% His efforts were crucial in managing the annotation process and ensuring high-quality data collection.

% \newpage
% \section*{Broader Impact}\label{sec:impact}
% \input{Arxiv/chapters/impact}

\bibliographystyle{plainnat}
\bibliography{bib.bib}

\clearpage

% %%%%%%%%%%%%%%%%%%%%%%%%%%%%%%%%%%%%%%%%%%%%%%%%%%%%%%%%%%%%
% \section*{Checklist}
% \input{Arxiv/chapters/checklist}

\newpage
\appendix

\section{Author Contributions}\label{sec:contributions}

\begin{itemize}
\item \textbf{Arshia Hemmat:} {
\begin{itemize}
    \item \emph{Dataset Generation:} Responsible for the development of the Illusion Generation Pipeline, meticulously designing and implementing the system to create high-quality illusions. Efforts included fine-tuning the hyperparameters to ensure the generated images met the desired standards of clarity and effectiveness. This process involved extensive experimentation and adjustment to achieve optimal results.
    
    \item \emph{Experiments and Method:} Conducted the zero-shot experiments for all the models, which involved setting up, running, and analyzing the results of these experiments.
    
    \item \emph{Data Annotation:} Managed data annotation involving over 100 individuals, collected and analyzed the annotation results to ensure data quality and relevance.
    
    \item \emph{Paper Writing:} Co-writer of the basics of the chapter \cref{sec:zs_exp}.
    
\end{itemize}
}
\item \textbf{Adam Davies:}
    \begin{itemize}
        \item \emph{Experiments and Method:} Co-supervised benchmark design, generation, evaluation, zero-shot experiments, and results analysis/visualization; implemented experimental prototypes for VLM generation and evaluation; assisted in model selection and deployment.
        \item \emph{Data Annotation:} Designed data annotation procedure; wrote onboarding materials for annotators; prototyped data annotation setup and assisted in deployment and analysis.
        \item \emph{Paper Writing:} Co-wrote, edited, and revised all sections of the paper; co-shaped central story (motivation, contribution, relationship with prior work in computer and human vision, results analysis, social impact); literature review (shape recognition in human vision, shape recognition benchmarks).
        % co-wrote 
        % Abstract, Introduction (\cref{sec:intro}), Related Work (\cref{sec:rel_works}) Benchmark Description (\cref{sec:benchmark}), Zero-Shot Experiments (\cref{sec:zs_exp}), Societal Impact (\cref{sec:impact}), Conclusion (\cref{sec:conclusion}), and \cref{app:extra_data_details,app:fullzeroshot,app:misc};
        % % introduction, related work, benchmark description, societal impact, conclusion, and appendices; 
        % edited and revised all sections of main paper and appendix.
    \end{itemize}
\item \textbf{Tom A. Lamb:}  \begin{itemize}
        \item \emph{Experiments:} Co-shaped design of ICL experiments; implemented and carried out all ICL experiments and co-led the analysis and presentation of ICL results.
        \item \emph{Paper Writing:} Led writing of \cref{sec:icl_exp}. Additionally, contributed to  the writing and presentation of \cref{sec:zs_exp}. 
        % introduction, related work, benchmark description, societal impact, conclusion, and appendices; 
    \end{itemize} 
\item \textbf{Jianhao Yuan:} 
\begin{itemize}
        \item \emph{Experiments:} Carried out all domain generalization experiments \cref{sec:mdg_exp}. Additionally, carried out zeroshot experiment of 
        % GPT4o \cite{openai2023gpt4} and Gemini \cite{team2023gemini} 
        GPT-4o and Gemini in \cref{sec:zs_exp}.  
        \item \emph{Paper Writing:} Led writing of \cref{sec:mdg_exp}. 
    \end{itemize} 
\item \textbf{Philip Torr:} Provided feedback and advice regarding the project direction and proposed approach; assisted in securing compute resources to carry out experiments.
\item \textbf{Ashkan Khakzar:} 
\begin{itemize}
\item \emph{Idea} Conceived the research problem and idea (the idea to evaluate VLMs on shapes represented by an arrangement of visual scene elements)
\item \emph{Method} Identified the existing method to generate such images. Demonstrated proof of concept (that state-of-the-art VLMs cannot identify these shapes)
\item \emph{Literature review} On shape recognition in computer vision and co-shaped the storyline
\item \emph{Experiments} Project co-supervision (curating the dataset \cref{sec:benchmark}, and zero-shot experiments \cref{sec:zs_exp}) 
\item \emph{Writing} Co-writing of abstract, introduction, related works, and conclusion.
\end{itemize}
\item \textbf{Francesco Pinto:} 
    \begin{itemize}
        \item \emph{Experiments and Method:} Led benchmark design, generation and evaluation; led design, results analysis and visualization of zero-shot, in-context learning and multi-domain generalization experiments.
        \item \emph{Data Annotation:} Implemented and tested data annotation procedure, assisted in preparing the onboarding materials for annotators; collected and analysed the results of the annotation.
        \item \emph{Paper Writing:} Co-shaped central story (motivation, contribution, relationship with prior work in computer vision, results analysis, impact); literature review (on shape recognition in computer vision); co-wrote the abstract and all sections of the paper but conclusions. Co-supervised full project.
        % introduction, related work, benchmark description, societal impact, conclusion, and appendices; 
    \end{itemize}
\end{itemize}

\clearpage

% \section{Appendix}
\section{Dataset Documentation and Additional Information}
\label{app:extra_data_details}

Below, we include all information required for dataset submissions to the NeurIPS Datasets and Benchmarks Track:

\paragraph{Dataset Documentation and Intended Uses} The dataset documentation is provided at the Croissant and Huggingface URLs mentioned below. The dataset mainly evaluates foundational VLMs and their shape recognition abilities. The dataset can also learn invariant representations using domain generalisation techniques. Other uses may be possible.  

\paragraph{Dataset URL} 
Our datasets are available for viewing and full download at the following permanent link: \url{https://huggingface.co/datasets/arshiahemmat/IllusionBench}.
The ``dataset viewer'' allows one to select a specific split (i.e., \sindata, \logos, or \icons). All images are provided in the \texttt{.png} format.
The HuggingFace Datasets repository service (where our dataset is hosted) automatically generates structured Web standard metadata for dataset discovery.
% {\red TODO: must confirm that HuggingFace Datasets actually does this (see \url{https://neurips.cc/Conferences/2024/CallForDatasetsBenchmarks#:~:text=Add%20structured%20metadata}) -- once confirmed, add URL link here}

\paragraph{Croissant Metadata URL} 
Our Croissant metadata record is available at \url{https://huggingface.co/api/datasets/arshiahemmat/IllusionBench/croissant}.

\paragraph{Author Statement} The authors have collected the conditioning images and generated this dataset for research purposes. For this reason, the data usage is allowed under the fair use law and is not intended to yield any copyright infringement. There is no warranty of fitness for a particular purpose or noninfringement. The authors remain available to edit the dataset to comply with the law. In no event shall the authors or the NeurIPS conference be liable for any claim, damages, or other liability arising from, out of, or in connection with the usage or release of this dataset.

\paragraph{Data License}
This work is openly licensed under CC BY-NC 4.0 (\url{https://creativecommons.org/licenses/by-nc/4.0/deed.en}).

% {\red TODO -- consult \url{https://paperswithcode.com/datasets/license} for options; I'm going to add a template for this but I don't really know which option is best, so please -- anyone who has experience with or knowledge about this, replace this license with whatever you think is best!}

\paragraph{Long-Term Hosting, Licensing, and Maintenance Plan}
We have uploaded our dataset to HuggingFace Datasets (link above). The Licensing information and Croissant metadata URL are available above and also available in the HuggingFace URL. Regarding Maintenance of the dataset on the HuggingFace servers please refer to the \url{https://huggingface.co/content-guidelines}.

\paragraph{Reproducibility}
The code for generating the dataset and the experiments are publicly available in the following repository \url{https://github.com/arshiahemmat/IllusionBench}.

\paragraph{Human Annotations}
We have provided screenshots of annotation forms which were distributed among participants in \cref{app:annotation}.
% Moreover, the onboarding instructions are provided in the supplementary.

\paragraph{Attributions}
This work utilizes stock images to condition generators (as described in \cref{sec:benchmark}). \icons conditioning images are taken from \url{icons8.com}, which makes them freely available provided they are attributed using a link (as we do here). 
% {\red TODO: give attribution for \sindata and \logos, including WWF panda (see \url{https://help.worldwildlife.org/hc/en-us/articles/360007905374-WWF-Panda-Logo}}

\subsection{Human Annotation Details}
\label{app:annotation}

\paragraph{Subsampling for Annotation} Given the size of our dataset ( more than 32K samples) performing a complete annotation of it would be expensive. Furthermore, since the data is synthesized and we perfectly know the class of the shapes represented in each image, the purpose of the annotation is simply to verify that the generated images have shapes that are recognisable by humans. For this reason, we subsample the generated dataset by enforcing that, for each dataset (i.e., each of \sindata, \icons, and \logos),
at least one conditioning image from each class and scene choice is annotated.

Furthermore, we observe that the difficulty in perceiving an object depends on the choice of the hyperparameters that control the diffusion process. For this reason, we additionally enforce that images are uniformly sampled from each hyperparameter setting so that annotators are exposed to images encompassing the full range of difficulty. 
%By doing so, we created a dataset with full coverage of the original dataset for human evaluation, ensuring that our human annotation dataset accurately represents the diversity and complexity of the entire dataset.

%Our solution was to select at least one conditioning image from each class in each dataset. In Table [?], we can see that several conditioning images were taken for each class to create the human annotation dataset. 

%It was important for us to accurately measure the standard of our dataset, so we distributed the difficulty of the images uniformly and selected images with varying levels of difficulty to ensure comprehensive coverage of the entire dataset.
%We ensured that each class in the main dataset was represented in the human annotation dataset by including a conditioning image from each class. Given that the scenes chosen are different and distinct, we also included all eleven scenes from the main dataset in the human dataset.

\paragraph{Participants}
Our human evaluation involved 106 participants. The annotators were first instructed about the task and required to perform a simple test on 10 images, in order to make sure they understood the task to be performed. 
Annotators participated on a purely volunteer basis and were awarded with in-course credit. Participation was not mandatory for any student or course. No risks were identified for the annotation process.

\paragraph{Further annotation meta-data} The number of annotators for \sindata, \icons, and \logos is respectively of 35 each. We split the original data so that each sample is annotated twice. Each reviewer is assigned approximately 80 samples (since the splitting algorithm is randomised, they may receive slightly less or slightly more samples).

\begin{figure}
    \centering
    \includegraphics[width=0.7\linewidth]{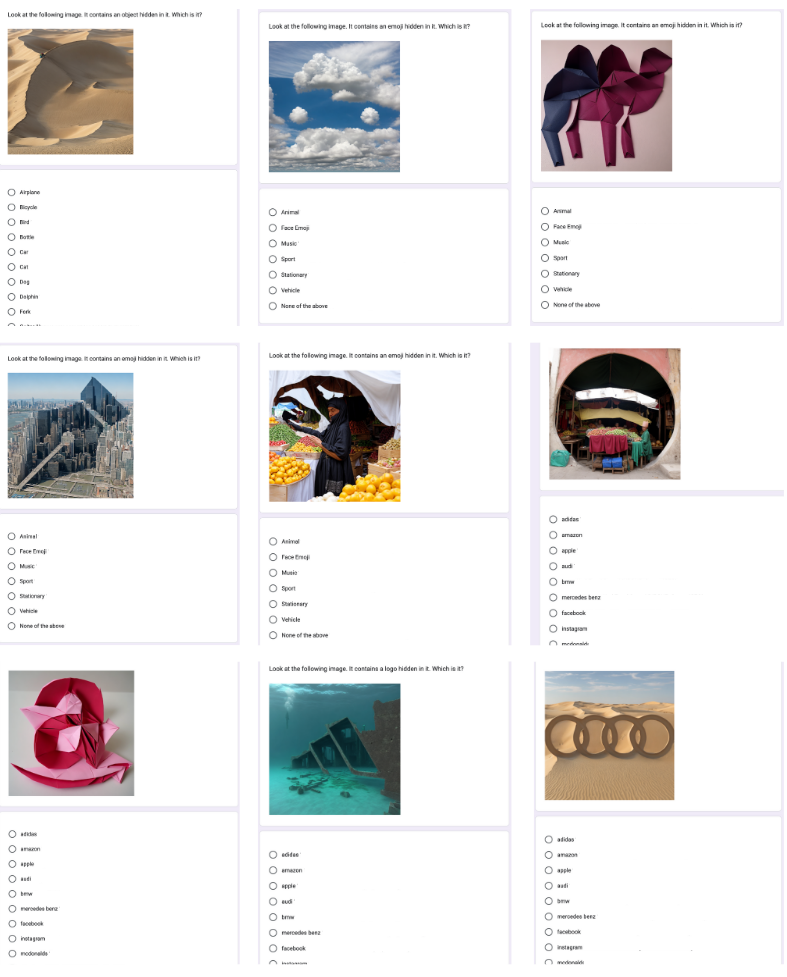}
    \caption{\textbf{Human Annotation} screenshots from GoogleForm through which the images are annotated by human annotators.}
    \label{fig:annotationForms}
\end{figure}
% {\red TODO: give age stats (min, max, mean, std dev)}

% \fp{I CAN SEE THIS BEING PROBLEMATIC: Course credit was awarded only to those participants who achieved an accuracy rate of more than 90\% in recording their answers. This ensured a high standard of data quality and participant engagement throughout the annotation process.}
% {\red ADAM: strongly agreed. this needs to go.}
%the test either online or in person. Given that data annotation requires prior knowledge about various tasks and the testing procedure, we prepared an onboarding file. This file was provided to participants to familiarize them with the method before they began the annotation process.

%Participants were motivated to join the experiment through two main incentives: course credit and voluntary participation.

\subsection{Image Generation Hyperparameters}
\label{sec:app_hyperparams}

% \paragraph{Image Generation Hyperparameters}
For data generation, we focused on the Illusion Diffusion generative models (demo available \href{https://huggingface.co/spaces/AP123/IllusionDiffusion}{here}), containing three major components:

\begin{itemize}
    \item ControlNet \cite{zhang2023addingControlNet}, specifically: \href{https://huggingface.co/monster-labs/control_v1p_sd15_qrcode_monster}{controlv1p sd15 qrcode monster}
    \item Base Model, specifically: \href{https://huggingface.co/SG161222/Realistic_Vision_V5.1_noVAE}{RealisticVision V5.1 noVAE}, built using Stable Diffusion \cite{rombach2022highSD}
    \item Stable Diffusion-guided VAE, specifically: \href{https://huggingface.co/stabilityai/sd-vae-ft-mse}{sd-vae-ft-mse}
\end{itemize}

% For the image generation, we have these parameters:
% \begin{itemize}
%     \item Conditioning Image
%     \item Prompts
%     \item Negative Prompts
%     \item Guidance-scale
%     \item  illusion\_strength
%     \item start of ControlNet
%     \item end of ControlNet
%     \item strength of the upscaler
%     \item sampler
% \end{itemize}

% \paragraph{Data Generation Setting}
We used the following generation hyperparameters:

\begin{itemize}
    % \item Conditioning Image: We gonna talk about the Conditioning images later but we downloaded our images from \href{https://icons8.com}{ICON8} and just we downloaded \textbf{Free Images}.
    \item Prompts were simply a single word corresponding to the scene types (e.g., ``city'' or ``museum'')
    % \item Negative Prompts: Always set to "low quality"
    \item Guidance-scale was always set to default value \textbf{7.5}
    \item Illusion\_strength, which can be used to modulate the strength of abstract shape patterns, was selected based on our anecdotal observations regarding an appropriate difficulty level for each dataset (see below) and validated using human data annotation (as described above)
    % datasets, we chose 13 values for the \logos and \sindata and 4 values for the \icons
    % \item start of ControlNet: Always set to zero
    % \item end of ControlNet: Always set to one
    % \item strength of the upscaler: Always set to zero
    \item Sampler was always set to default value \textbf{Euler}
\end{itemize}

\begin{tcolorbox}
The Illusion\_strength for the different datasets are as follows:\\
\begin{itemize}
    \item \{Illusion\_strength\} of the \logos and \sindata: [0.75, 0.80, 0.85, 0.90, 1.05, 1.10, 1.15, 1.20, 1.25, 1.35, 1.40, 1.50, 1.60]
    \\
    \item \{Illusion\_strength\} of the \icons:  [0.85, 1.05, 1.25, 1.40]
\end{itemize}
\end{tcolorbox}

% MUST KEEP THIS SECTION!!! WE REFERENCE IT IN THE CHECKLIST!!!
\subsection{Limitations}
\label{app:dataset_limitations}
For future work, we will create more complex images and define more tasks in order to challenge models. We have also increased the size of our dataset so that we can train large models using our dataset. A current limitation is that we only hide a single shape in each image. Future work could extend this to incorporating several objects within the same background. Finally, we also plan to experiment with further tasks for compositional understanding and scene understanding of SOTA models. We leveraged prompt engineering to report the best possible performance of each model in the zero-shot case as described in \cref{app:fullzeroshot} and \cref{sec:zs_exp}, however, improvements may be possible. We describe several limitations of the methods explored in this work in \cref{sec:zs_exp,sec:icl_exp}.
% MUST KEEP THIS SECTION!!! WE REFERENCE IT IN THE CHECKLIST!!!

\subsection{Data Samples}
\begin{figure}[ht!]
  \centering
  % First image
  \begin{minipage}{0.45\textwidth}
    \centering
    \includegraphics[width=\linewidth]{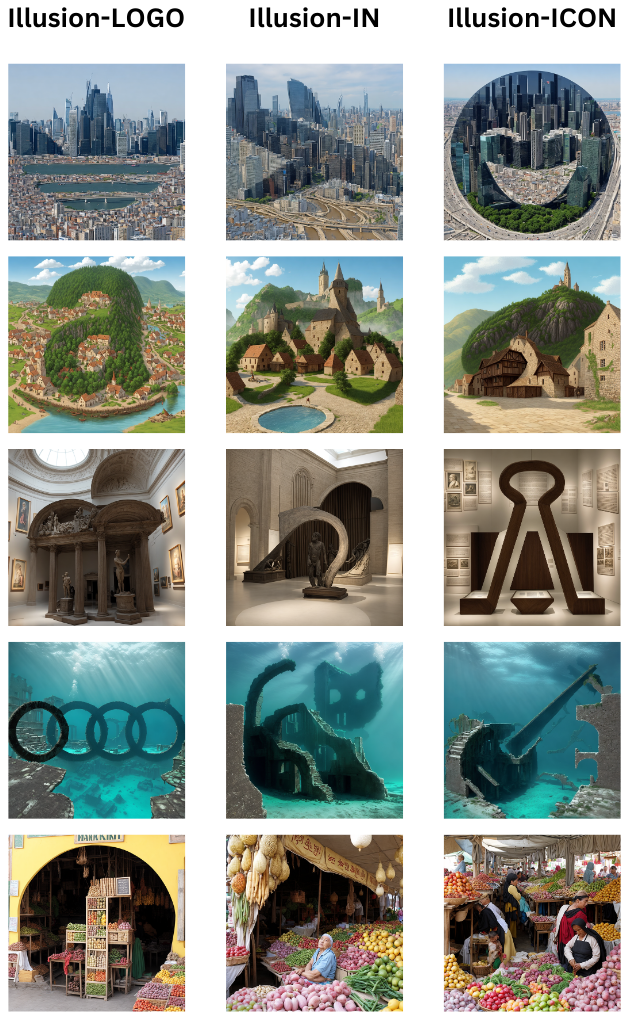} 
  \end{minipage}
  \hfill
  % Second image
  \begin{minipage}{0.45\textwidth}
    \centering
    \includegraphics[width=\linewidth]{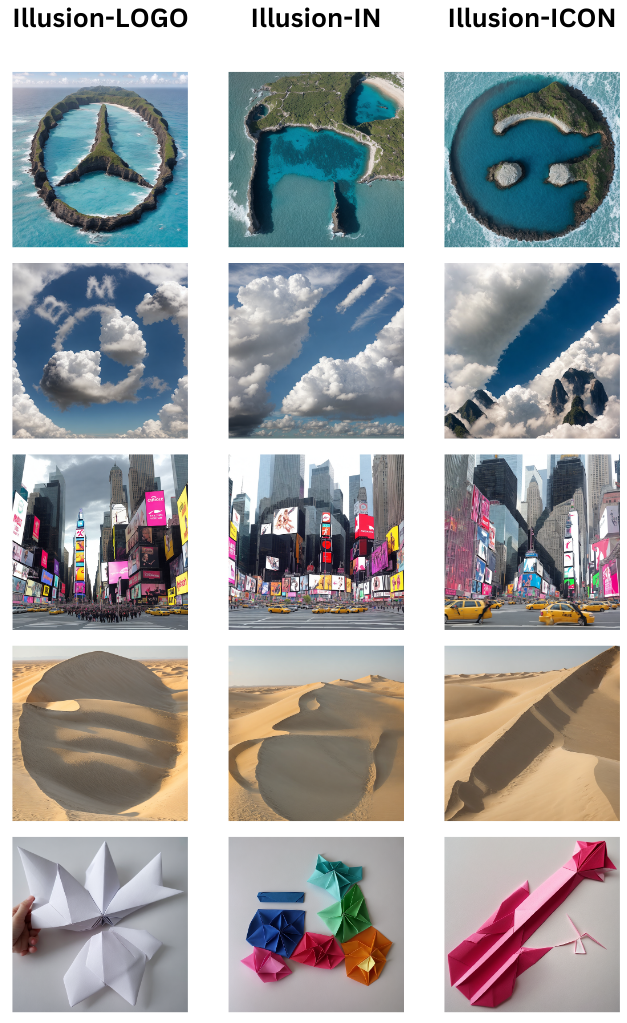} % adjust the filename and width as needed
  \end{minipage}

  \caption{\textbf{Image Samples} from each dataset in our benchmark.} % Single caption for both images
    \label{fig:samples}
\end{figure}

To illustrate the quality of abstract shape recognition images created for this dataset, we randomly sample one image from several scene types in each dataset and display them in \cref{fig:samples}.

\section{Zero-Shot Experiments Details}
\label{app:fullzeroshot}

\subsection{Zero-shot Experiments}
We test our models zero-shot to evaluate their abstract shape recognition abilities. To leverage all capabilities of these models, we describe the conditions of our experiments in our prompt. The models are then asked to choose the correct shape type among a closed set of options, which include both shapes and scene names.

Let us focus on the predictive task $\tc$. Analogous formulations hold for $\ts$ and $\tb$. Given that the model can correctly assign the class $\yc{i_{*}}$ to the hidden shape in the scene $\xc{i_{*}}$, we provide it with a set of options $\mathcal{O}$, which includes all the shapes and scene names considered in the dataset split. We then ask the model to predict the shape name from these options.

Define $\mathcal{O} = \{\text{shape}_1, \text{shape}_2, \ldots, \text{scene}_1, \text{scene}_2, \ldots \}$ as the set of possible options. The model's response is evaluated based on whether the correct shape name is present in its output.

% {\red TODO: add information about the generation hyperparameters (e.g., length normalization, temperature, random sampling (such as top-k or nucleus) if applicable, etc.)}

\subsection{Models}
\label{sec:app_vlm_models}

In our zero-shot experiments, we evaluate each of the following large vision language models (VLMs):
\begin{itemize}
    \item BlipV2-T5 \citep{li2023blip2}, a VLM utilizing the T5 architecture \citep{raffel2020exploring} for text encoding and a state-of-the-art vision encoder, designed for high-performance multimodal tasks.
    \item CogVLM \citep{wang2024cogvlm}, an advanced VLM leveraging a Vision Transformer (ViT) \citep{dosovitskiy2021image} and a powerful language model fine-tuned for vision-language reasoning tasks.
    \item InstructBlip-T5 \citep{dai2023instructblip}, a model combining the T5 architecture \citep{raffel2020exploring} for text processing with a highly efficient vision encoder, fine-tuned for instructional prompts and multimodal interactions.
    \item LLaVA-Next (Vicuna-7b) \citep{liu2024llavanext}, a VLM\
    % operating at an input image resolution of $336^2$, 
    using Vicuna-7b-v1.5 \citep{zheng2024judging} and CLIP ViT-L/14 \citep{radford2021learning} as text and visual encoders, respectively. These are connected via simple projections.
    \item Qwen-VL-Chat \citep{bai2023qwen}, a 9B parameter model 
    % with an input resolution of $448^2$, 
    employing a cross-attention module to link an OpenClip ViT-bigG \citep{ilharco_2021_5143773} vision encoder to a Qwen-7b \citep{bai2023qwen} text backbone.
    \item Llava1.5-7b and 13-b \citep{liu2024improved}, a VLM 
    % operating at an input image resolution of $336^2$, 
    employing a 7-billion parameter language model and advanced visual encoder, connected via efficient projections.
    \item InstructBlip-7b and 13b \cite{dai2023instructblip,dai2024instructblip}, a BLIP \cite{li2022blip} model fine-tuned using instruction tuning, using a 7-billion parameter language model and a high-resolution vision encoder for precise multimodal understanding.
    \item MoE-StableLM, MoE-Qwen, MoE-Phi2 \citep{lin2024moellava}, a mixture of experts (MoE) model combining StableLM architecture \citep{raffel2020exploring} with multiple expert models for dynamic task specialization and improved performance. 
    \item GPT-4o, a multimodal version of GPT-4 \citep{openai2023gpt4}, incorporating optimized end-to-end multimodal encoding of images, text, and audio for improved multimodal task performance.
    \item Gemini-Flash \citep{team2023gemini}, a high-speed VLM combining the latest advancements in vision transformers \citep{dosovitskiy2021image} and language models for rapid and accurate multimodal analysis.
\end{itemize}
Note that, for the last two models in this list, we are unable to provide any specific information regarding their respective architectures or training regimes, as this information has not been made publicly available.

\subsection{Prompts}
\label{app:zsl_prompts}

We use the following general prompt template for our zero-shot experiments:

\begin{tcolorbox}
    \begin{itemize}
        \item T1 Prompt: \texttt{This image contains a \{shape\} integrated into a background, where elements of the background contribute to forming the \{shape\}. Identify the \{shape\} that is represented in the image by choosing exclusively among the following options: \{shape\_options\}, \{background\_classes\}. Provide your response by stating only the single, most accurate class name that represents the \{shape\}. You have to respond with a single word.}

        \item Texture Question Bias: \texttt{This image contains a \{shape\} integrated into a background, where elements of the background contribute to forming the \{shape\}. Identify the background that is represented in the image by choosing exclusively among the following options: \{shape\_options\}, \{background\_classes\}. Provide your response by stating only the single, most accurate class name that represents the background. You have to respond with a single word.}
        
    \end{itemize}
\end{tcolorbox}

where \texttt{shape} $\in \{$logo, shape, icon$\}$for the dataset IllusionBench-LOGO, IllusionBench-IN and IllusionBench-CI respectively.

\subsection{Text Generation Hyperparameters}
\label{sec:vlm_hypers_zs}
For all VLMs, we use full-precision weights (i.e., no quantization), generating responses using greedy decoding without sampling, and limit the maximum response length to 100 tokens.

\newpage

% \subsection{Individual Dataset Splits Zero-shot performance (compared Stylized ImageNet)}
\subsection{Zero-Shot Results By Class}
\label{app:individual_dataset_results_zeroshot}

\begin{figure}[h!]
    \centering
    \includegraphics[width=\textwidth]{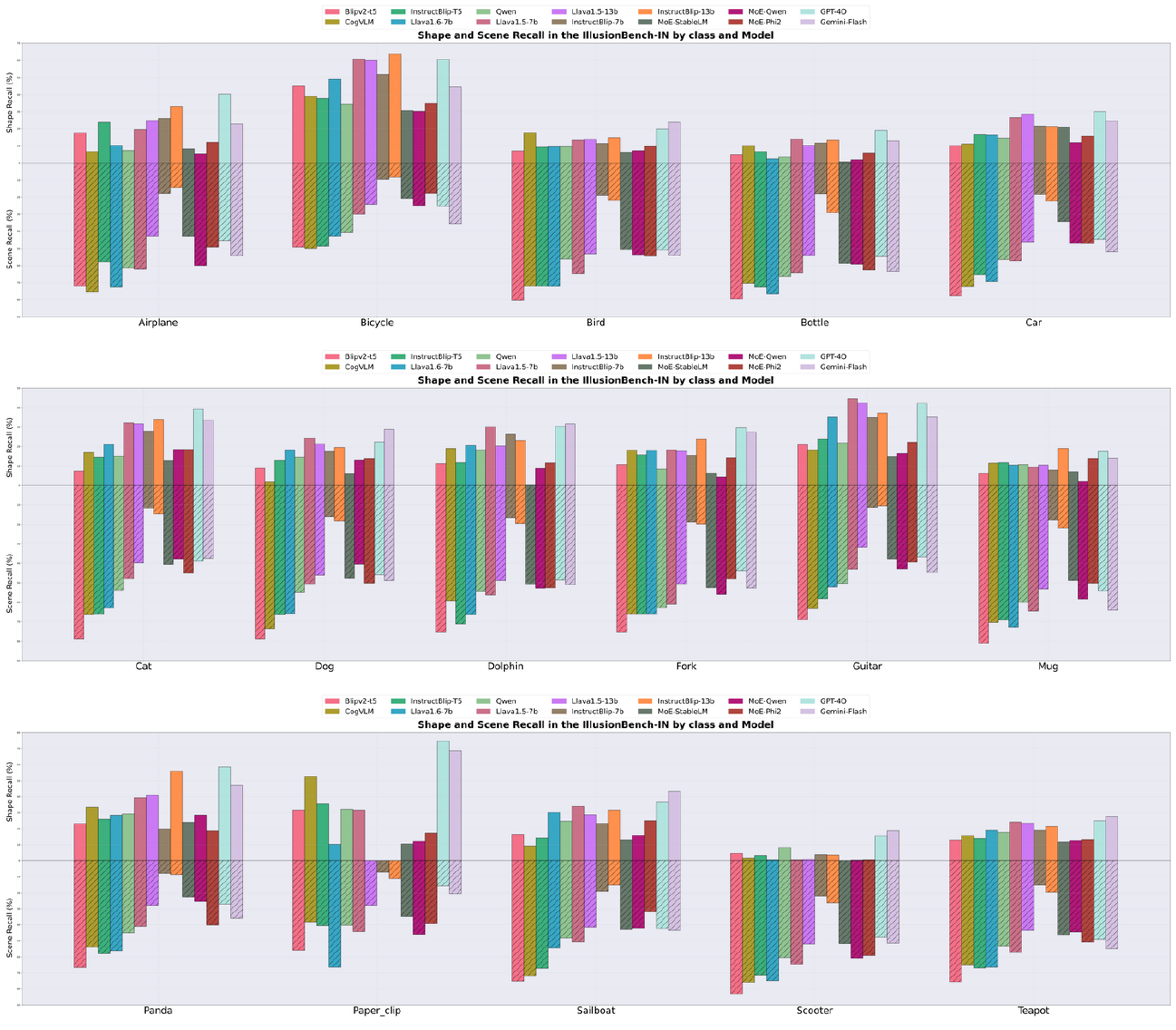}
    \caption{\textbf{Zero-shot results on \sindata by class.} Zero-shot shape and scene recall of VLMs for each class in the \sindata dataset.}
    \label{fig:zsl_sinp}
\end{figure}

\clearpage

\begin{figure}[ht!]
    \centering
    \includegraphics[width=\textwidth]{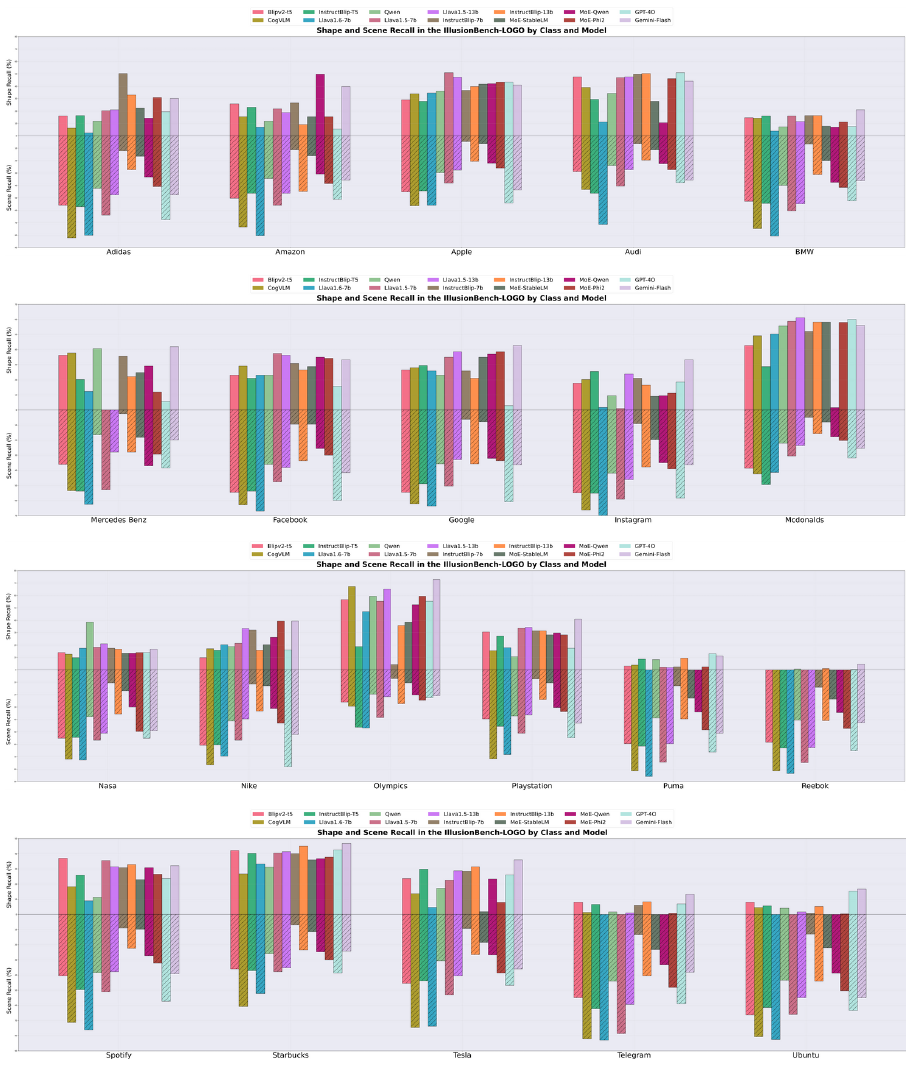}
    \caption{\textbf{Zero-shot results on \logos by class.} Zero-shot shape and scene recall of VLMs for each class in the \logos dataset.}
    \label{fig:zsl_logos}
\end{figure}

\begin{figure}[ht!]
    \centering
    \includegraphics[width=\textwidth]{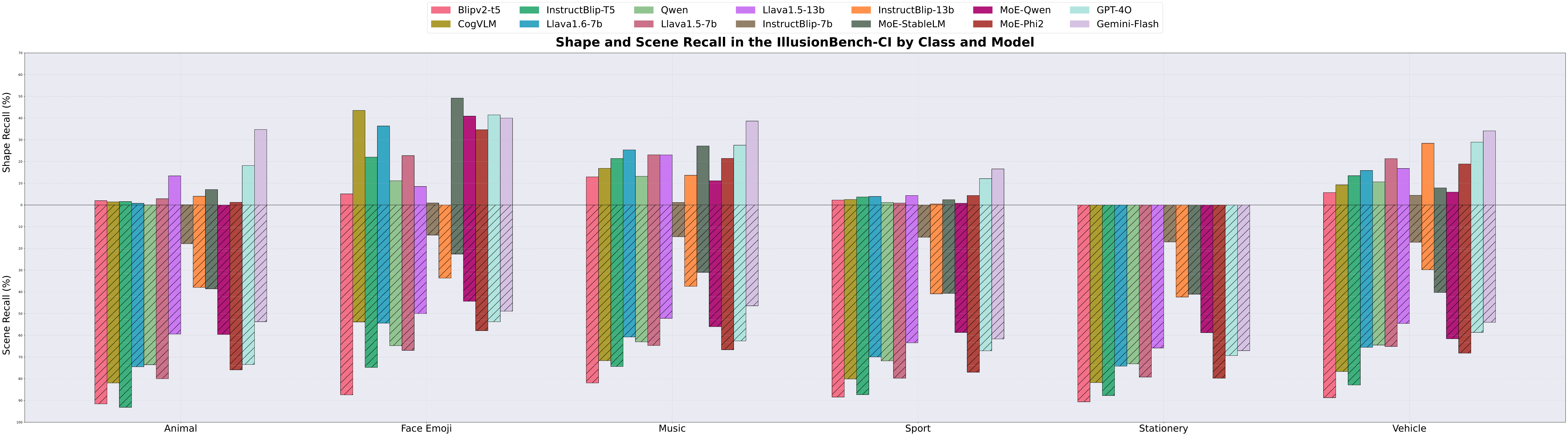}
    \caption{\textbf{Zero-shot results on \icons by class.} Zero-shot shape and scene recall of VLMs for each class in the \icons dataset.}
    \label{fig:zsl_icons}
\end{figure}

\begin{figure}[ht!]
    \centering
    \includegraphics[width=\textwidth]{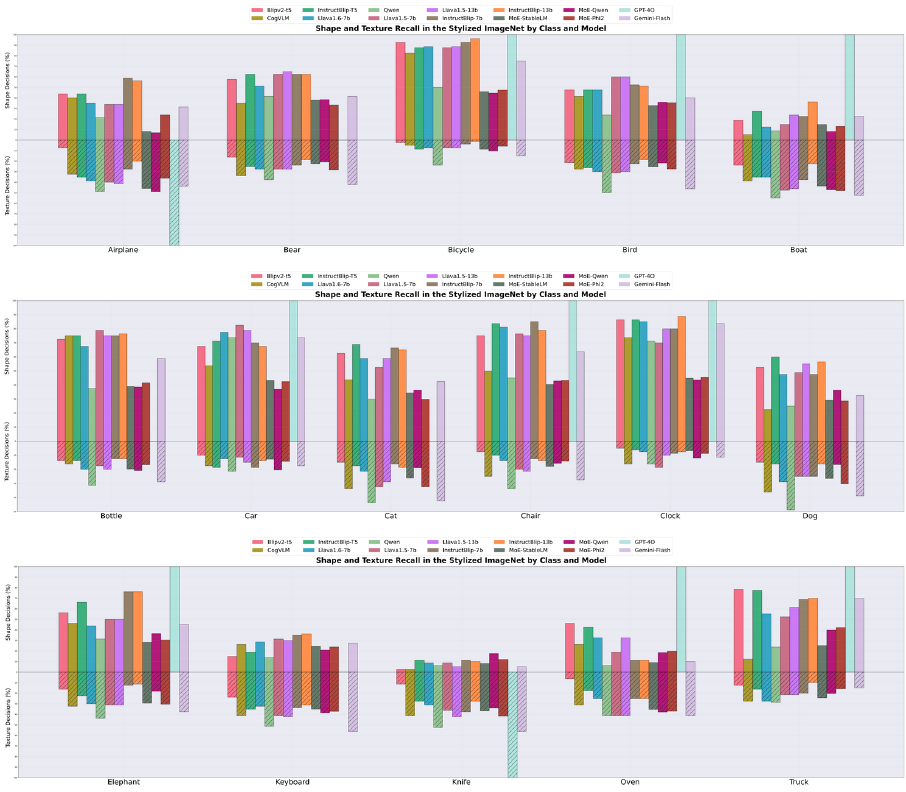}
    \caption{\textbf{Zero-shot results on Stylized ImageNet by class.} For comparison, we also report the zero-shot shape and texture bias of VLMs on the Stylized ImageNet dataset \cite{geirhos2018imagenettrained}.}
    \label{fig:zsl_sin}
\end{figure}

\pagebreak

\section{In-Context Learning Experiments Details}
\label{app:icl_full}
\subsection{In-Context Learning (ICL)}
\label{app:icl-description}
ICL is a method of adapting a model for an unseen task without any additional training or fine-tuning. Specifically, $n$-shot ICL consists of sequence of labelled demonstrations $\mathcal{C} = \{(x_{i_1}, y_{i_1}), \cdots , (x_{i_n}, y_{i_n}) \}$. These are supplied to a model $p_{\bm \theta}(y|x)$ for an unseen task. The label corresponding to a test query $x_*$ is predicted through the predictive distribution of the model conditioned on the demonstration set $\mathcal{C}$ alongside an instruction $I$ for the new task:
\begin{equation}
    p_{\bm \theta}(y|\mathcal{C}, I) = p_{\bm \theta}(y|x_{i_1}, y_{i_1}, \cdots x_{i_n}, y_{i_n}, I).
\end{equation}
 This learning method has proven to be an efficient and low-cost method for adapting LLMs to downstream tasks \citep{brown2020language, schick2021s, winata2021language, liu2022makes}. The success of ICL for LLMs has led to recent research aiming to extend ICL to multi-modal models, where labeled demonstrations now contain interleaved image and text modalities \citep{alayrac2022flamingo, bertini2024makes, zhao2023mmicl, zong2024vl}. 
\subsection{ICL Further Experimental Details}
\label{app:icl-experimental-details}
Considering we restrict evaluations to classes recognised in a zero-shot manner, we use the following class counts: 10 for the \logos split, 14 for the \sindata split, and 6 for the icons split, utilizing all 11 scenes of the dataset. To overcome ICL biases like majority voting and recency bias, each shape and scene class is represented at most once within the context, with no repetitions, and new demonstrations are randomly sampled for each test sample. 

\subsection{Models Description}
In our zero-shot experiments, we evaluate each of the following large vision language models (VLMs):
\label{app:icl_models}
\begin{itemize}
    \item LLaVA-Next (Vicuna-7b) \citep{liu2024llavanext}, a VLM operating at an input image resolution of $336^2$, using Vicuna-7b-v1.5 \citep{zheng2024judging} and CLIP ViT-L/14 \citep{radford2021learning} as text and visual encoders, respectively. These are connected via simple projections.
\item Qwen-VL-Chat \citep{bai2023qwen}, a 9B parameter model with an input resolution of $448^2$, employing a cross-attention module to link an OpenClip ViT-bigG \citep{ilharco_2021_5143773} vision encoder to a Qwen-7b \citep{bai2023qwen} text backbone.
\item Otter-MPT \citep{li2023mimic}, a 9B parameter VLM based on the OpenFlamingo architecture \citep{awadalla2023openflamingo}, featuring an input image resolution of $224^2$ and utilizing LLaMA-7B \citep{touvron2023llama} and CLIP-ViT-L/14 as text and image backbones, respectively, connected through cross-attention.
\item IDEFICS-9B-Instruct \citep{laurenccon2024obelics}, an open-source reproduction of Flamingo \citep{alayrac2022flamingo}, with an input image resolution of $224^2$, using cross-attention transformer blocks to connect LLaMA and OpenClip text and image backbones.
\item MMICL-T5-XXL \citep{zhao2023mmicl}, a 12B parameter model that employs a Q-former \citep{li2023blip} to integrate language and image components within an InstructBlip-FLANT5-XXL \citep{dai2024instructblip} backbone. This model can handle complex prompts with interleaved text and images, allowing for text-image references through dummy demonstration tokens, and operates at an input image resolution of $224^2$.
\end{itemize}

\subsection{Prompts}
\label{app:icl_prompts}
We use the following general prompt template for our ICL experiments:
\begin{tcolorbox}
\{\texttt{TASK\_INSTRUCTION}\} \\
\{\texttt{demonstration\_image\_1}\} \\
Answer: \{\texttt{demonstration\_label\_1}\} \\
\{\texttt{demonstration\_image\_2}\} \\
Answer: \{\texttt{demonstration\_label\_2}\} \\
\vdots \\
\{\texttt{demonstration\_image\_n}\} \\
Answer: \{\texttt{demonstration\_label\_n}\} \\
\{\texttt{query\_image}\} \\
Answer: 
\end{tcolorbox}
where \texttt{demonstration\_image\_i} and \texttt{demonstration\_label\_i} refer to the image and label for the $i$th demonstration used as the context for predicting the answer for the query image \texttt{query\_image}. \texttt{TASK\_INSTRUCTION} is the instruction used based on the prediction target and the dataset. We used the following \texttt{TASK\_INSTRUCTION} prompts for predicting the shape, texture, and both the texture and shape simultaneously respectively:

\begin{tcolorbox}
    \# Predict shape \\
    \texttt{TASK\_INSTRUCTION =`This image contains a \{shape\} integrated into a background, where elements of the background contribute to forming the image.\\background options: [\{BG\_OPTIONS\}]\\ \{shape\} options: [\{SHAPE\_OPTIONS\}] \\ Identify the \{shape\} that is represented in the image by choosing among the provided options. Provide your response by stating only the single, most accurate option that represents the \{shape\} in the image. You have to respond with a single word.'} \\ \\
    \# Predict texture \\
    \texttt{TASK\_INSTRUCTION = `This image contains a \{shape\} integrated into a background, where elements of the background contribute to forming the image.\\background options: [\{BG\_OPTIONS\}]\\ \{shape\} options: [\{SHAPE\_OPTIONS\}] \\ Identify the background that is represented in the image by choosing among the provided options. Provide your response by stating only the single, most accurate option that represents the background in the image. You have to respond with a single word.'} \\ \\ 
    \# Predict both texture and shape \\
    \texttt{TASK\_INSTRUCTION = `This image contains a \{shape\} integrated into a background, where elements of the background contribute to forming the image.\\background options: [\{BG\_OPTIONS\}]\\ \{shape\} options: [\{SHAPE\_OPTIONS\}] \\  Identify BOTH the background AND the \{shape\} that are represented in the image by choosing among the provided options. Provide your response by stating only the single, most accurate options that represent the background and the \{shape\} in the image respectively. You have to respond with two words, one predicting the background and one predicting the \{shape\}'}
    
\end{tcolorbox}

where \texttt{shape} $\in \{$logo, object, icon$\}$for the dataset IllusionBench-LOGO, IllusionBench-IN and IllusionBench-CI respectively.

\subsection{Text Generation Hyperparameters}
\label{sec:vlm_hypers_icl}
For all VLMs, we use full-precision weights (i.e., no quantization), generating responses using greedy decoding without sampling, and limit the maximum response length to 100 tokens.

\subsection{ICL Results: Exceptions}
\label{app:ICL-exceptions}
We list the exceptions to the general treneds reported in \cref{sec:icl_exp}. We maintain the key takeaway headings and format in \cref{sec:icl_exp} and discuss key exceptions.
\begin{itemize}
\item \textit{ICL does not mitigate tendency to predict scene over shape.}
\texttt{LLaVA} on the task $\tc$ (along the first row) stands as an exception, where the model demonstrates low scene prediction accuracy and non-trivial performance shape accuracy on ICL2 and ICL4. 

\item  \textit{Context selection strategy effects prediction tasks differently}.
    \begin{itemize}
        \item $\tc$ : For \texttt{LLaVA}, including the shape through ICL2 or ICL4 for 1 or 2 shots leads to a significant performance increase over all other models. This is especially evident for 1-shot, where we see high shape accuracy values of ICL2: $97.9\%$ and ICL4: $99.9\%$. These high accuracy values indicate that the model exhibits a copying phenomenon \citep{bertini2024makes}, where for 1-shot, it simply copies the label from the ICL demonstration, which will have the same test label.
    
         \item $\ts$: \texttt{QWEN} shows an improvement in scene accuracy (for 4-shot, scene accuracies are ICL1: $51.4\%$ and ICL3: $88.1\%$) when the scene is included in the context. Additionally, \texttt{LLaVA} exhibits a similar copying phenomenon for scene prediction in ICL3 and ICL4 as discussed for $\tc$ but also shows some improvements over zero-shot for 2-shots.
    
         \item $\tb$: As an exception,  \texttt{OTTER} and \texttt{QWEN} show a general increase in scene accuracy on $\tb$ compared to $\ts$, while their shape accuracy remains similar to $\tc$. This suggests that predicting both shape and scene and including demonstrations with such predictions can help these models better disentangle scene from shape. Again, we observe the copying mechanisms in \texttt{LLaVA} described for $\tc$ and $\ts$.
    \end{itemize}

\end{itemize}

\subsection{Individual Dataset Splits ICL Results}
\label{app:individual_dataset_results}

\begin{figure}[ht!]
    \centering
    \includegraphics[width=\textwidth]{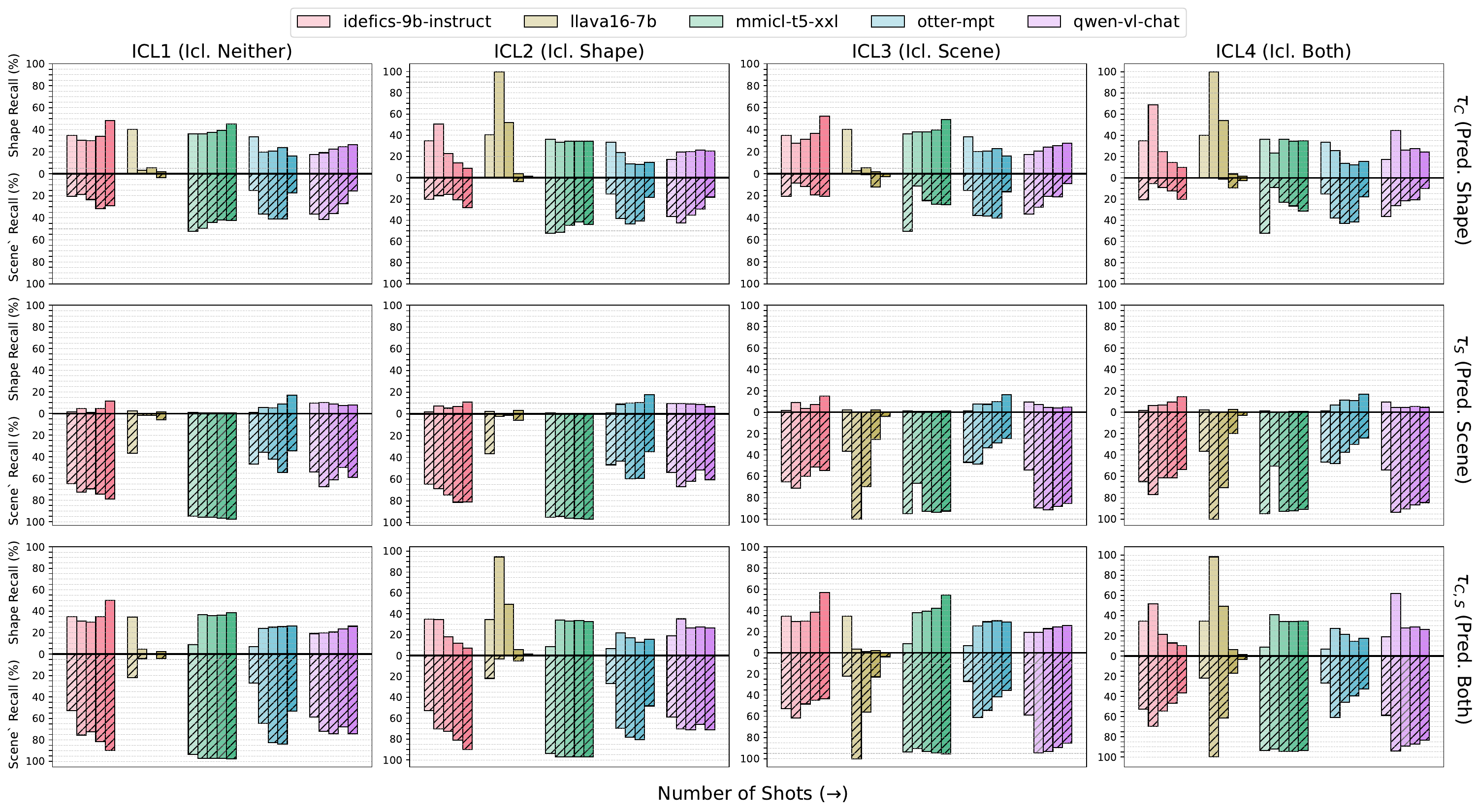}
    \caption{\textbf{ICL Results on \logos}. Few-shot shape and texture accuracy of VLMs on the \logos dataset across the different ICL learning tasks and the different prediction tasks. }
    \label{fig:icl_logos}
\end{figure}

\begin{figure}[ht!]
    \centering
    \includegraphics[width=\textwidth]{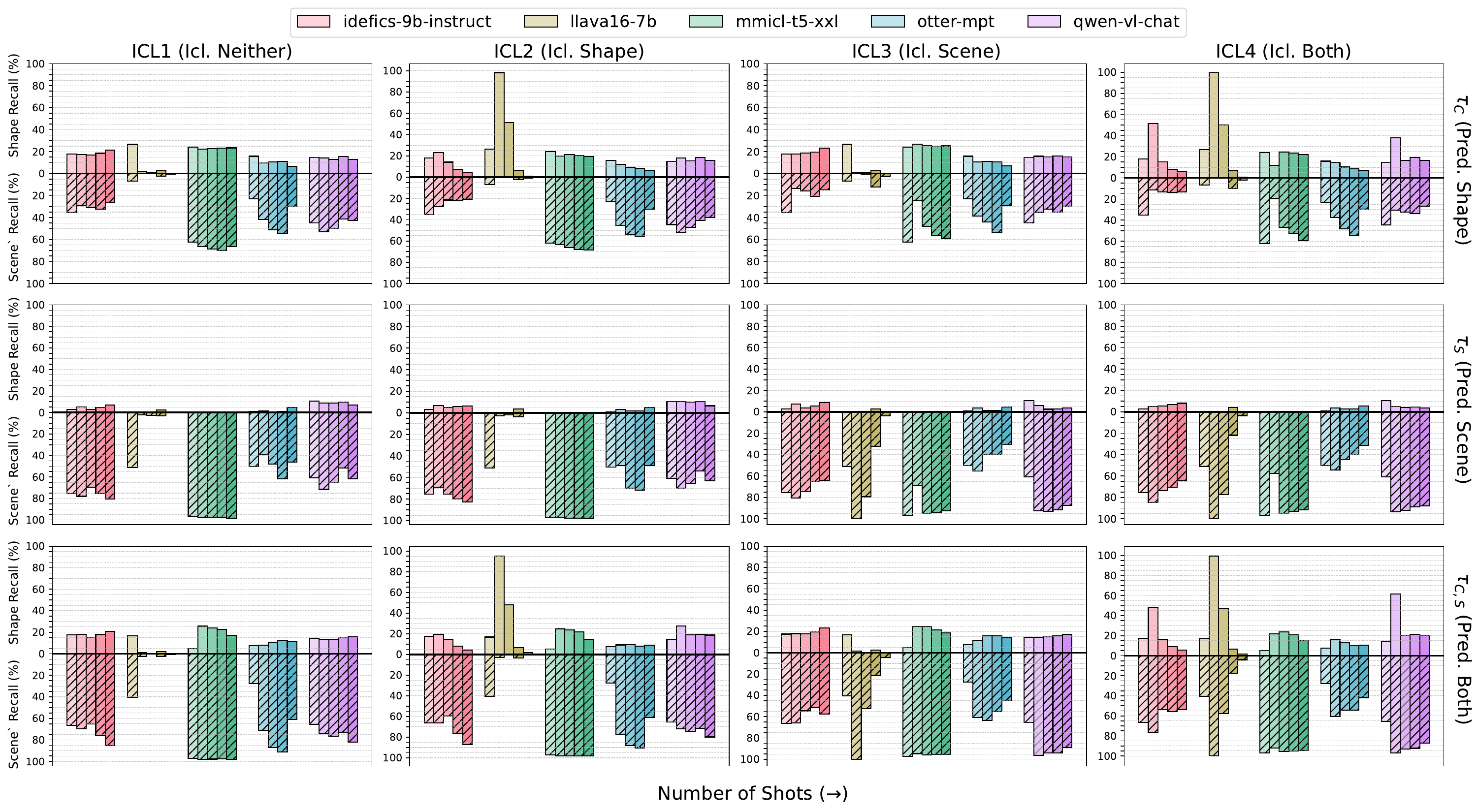}
    \caption{\textbf{ICL Results on \sindata}. Few-shot shape and texture accuracy of VLMs on the \sindata dataset across the different ICL learning tasks and the different prediction tasks. }
    \label{fig:icl_sin}
\end{figure}

\begin{figure}[ht!]
    \centering
    \includegraphics[width=\textwidth]{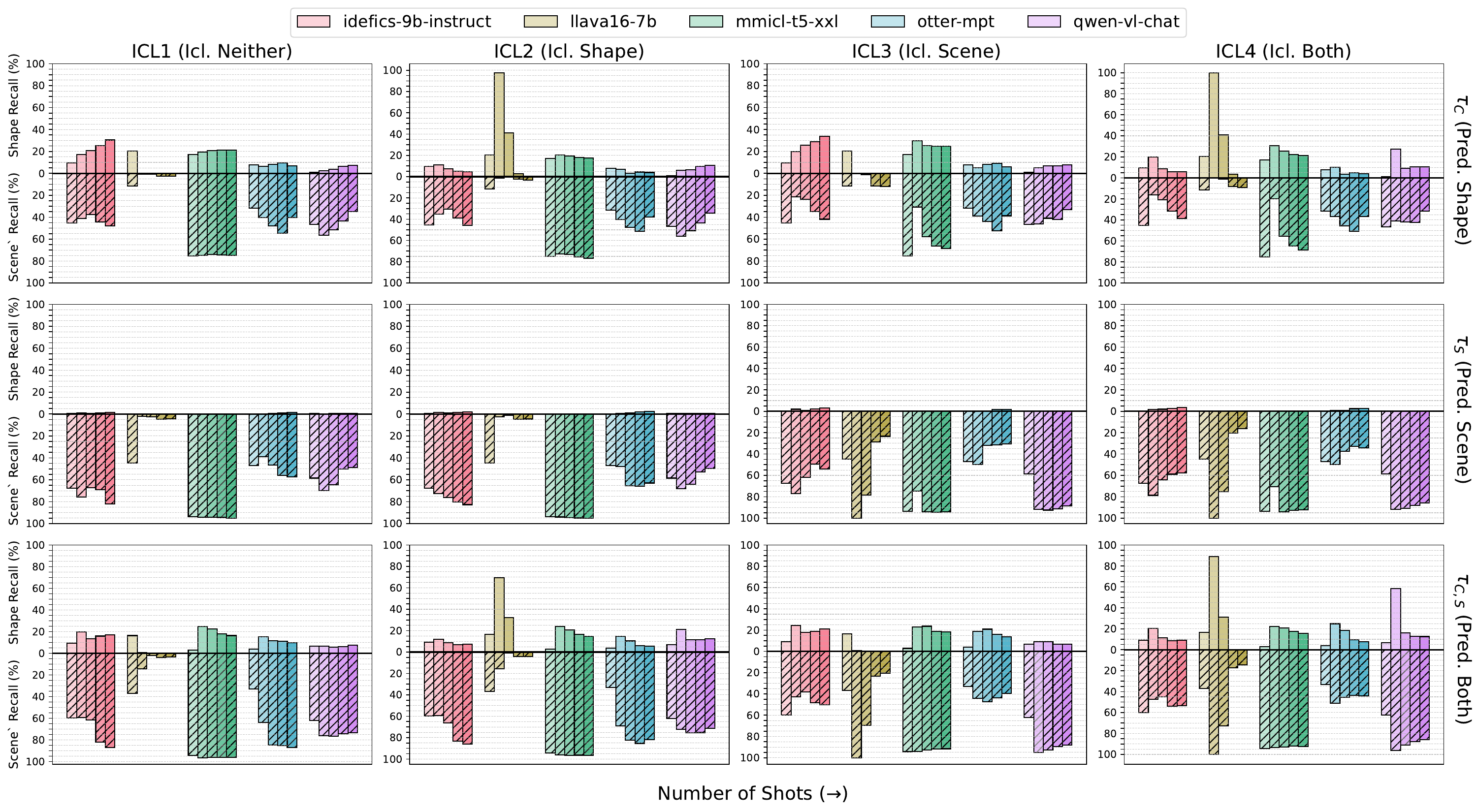}
    \caption{\textbf{ICL Results on \icons}. Few-shot shape and texture accuracy of VLMs on the \icons dataset across the different ICL learning tasks and the different prediction tasks. }
    \label{fig:icl_icons}
\end{figure}

\subsection{Responses From Low Performing Models}
\label{app: llava-bad-performance}
We often observe close to 0\% shape accuracy of the \texttt{LLaVA} model on shape prediction tasks across all four ICL-constrained ICL prediction tasks when using a higher number of ICL demonstrations. \Cref{fig:llava-verbose} illustrates three example responses from the \texttt{LLaVA} model using 4-shot ICL for ICL3, which includes the test query background in the ICL demonstrations. From the example model responses R1, R2, and R3, it is evident that the \texttt{LLaVA} model tends to produce descriptive and verbose responses. Specifically, it fails to be concise and accurate, unlike the other models we investigate that usually respond with a single class prediction even with more shots. This verbosity leads to poor accuracy as the model fails to adhere to the prompt instructions of predicting a single class, resulting in the test class rarely being included in the model's responses.

\begin{figure}[ht!]
    \begin{tcolorbox}
\begin{itemize}
    \item \textbf{R1}: \texttt{The image shows a paper sculpture that resembles a stylized}  
    \item \textbf{R2}: \texttt{The image shows a logo integrated into a background that features a mountainous landscape} 
    \item \textbf{R3}: \texttt{The image shows a beautiful natural scene with a large rock formation in the ocean}
\end{itemize}
\end{tcolorbox}
    \caption{\textbf{\texttt{LLaVA} verbose responses.} Example responses from the \texttt{LLaVA} model for 4-shot shape prediction (T1) on the ICL3 learning task.}
    \label{fig:llava-verbose}
\end{figure}

However, \cref{fig:llava-concise} shows example responses from the \texttt{LLaVA} model on the same task and for the same test queries as in \cref{fig:llava-verbose} but using 2-shots. Observations from responses R1', R2', and R3' indicate that with fewer shots, the model is much more likely to produce single-class predictions or responses that are generally more concise and less descriptive. The differences observed with increasing numbers of shots suggest that \texttt{LLaVA}'s ability to correctly process and learn both the expected answer format and the task diminishes with a greater number of shots, highlighting its limitation as an in-context learner.
\begin{figure}[ht!]
    \centering
\begin{tcolorbox}
\begin{itemize}
    \item \textbf{R1}': \texttt{The logo in the image is Tesla.}  
    \item \textbf{R2}': \texttt{The logo in the image is Starbucks.} 
    \item \textbf{R3}': \texttt{Audi}
\end{itemize}
\end{tcolorbox}
    \caption{\textbf{\texttt{LLaVA} concise responses.} Example responses from the \texttt{LLaVA} model for 2-shot shape prediction (T1) on the ICL3 learning task for the same test query as in \cref{fig:llava-verbose}.}
    \label{fig:llava-concise}
\end{figure}

\subsection{ICL Prompt and Context Sensitivity}
\label{app:ICL ablations}
\textbf{Prompt Sensitivity.} We assess whether our ICL results are sensitive to the prompts used. We conduct ablations over four different prompt templates: the original template provided in \Cref{app:icl_prompts} and three additional variations. These variations include: (i) a simplified minimalistic prompt, (ii) the same simplified prompt but reversing the order in which the object and background options are presented, and (iii) a Llama-guard-style prompt \citep{inan2023llama} that explicitly indicates what the model should and should not focus on when making predictions. The specific prompt templates are as follows:

\begin{tcolorbox}
\textbf{Simplified Prompt:}\\
\texttt{TASK\_INSTRUCTION = `This image contains an \{object\} integrated into a background, where elements of the background contribute to forming the image.\\
background options: [\{BG\_OPTIONS\}]\\
\{object\} options: [\{OBJ\_OPTIONS\}]\\
Identify the \{object/background/object and background\} that are represented in the image by choosing among the provided options.'}

\textbf{Simplified Prompt (Reverse):}\\
\texttt{TASK\_INSTRUCTION = `This image contains a background with an integrated \{object\}, where elements of the background contribute to forming the image.\\
\{object\} options: [\{OBJ\_OPTIONS\}]\\
background options: [\{BG\_OPTIONS\}]\\
Identify the \{object/background/object and background\} that are represented in the image by choosing among the provided options.'}

\textbf{Llama-Guard Style:}\\
\texttt{TASK\_INSTRUCTION = `This image contains an \{object\} integrated into a background, where elements of the background contribute to forming the image.\\
\{object\} options: [\{OBJ\_OPTIONS\}]\\
background options: [\{BG\_OPTIONS\}]\\
Identify the \{object/background/object and background\} that is represented in the image by choosing among the provided options. Provide your response by stating only the single, most accurate option that represents the \{object/background/object and background\} in the image. You have to respond with a single word.'}

\#\#\# \textbf{Pay attention to:}\\
\- ONLY \{the object/the background/BOTH the object and the background\} that is represented in the image by choosing among the provided icon options.

\#\#\# \textbf{DO NOT:}\\
\- Focus on the \{object/the background/IGNORE IN THIS CASE\} of the image.
\end{tcolorbox}

\begin{figure}
    \centering
    \includegraphics[width=1.0\linewidth]{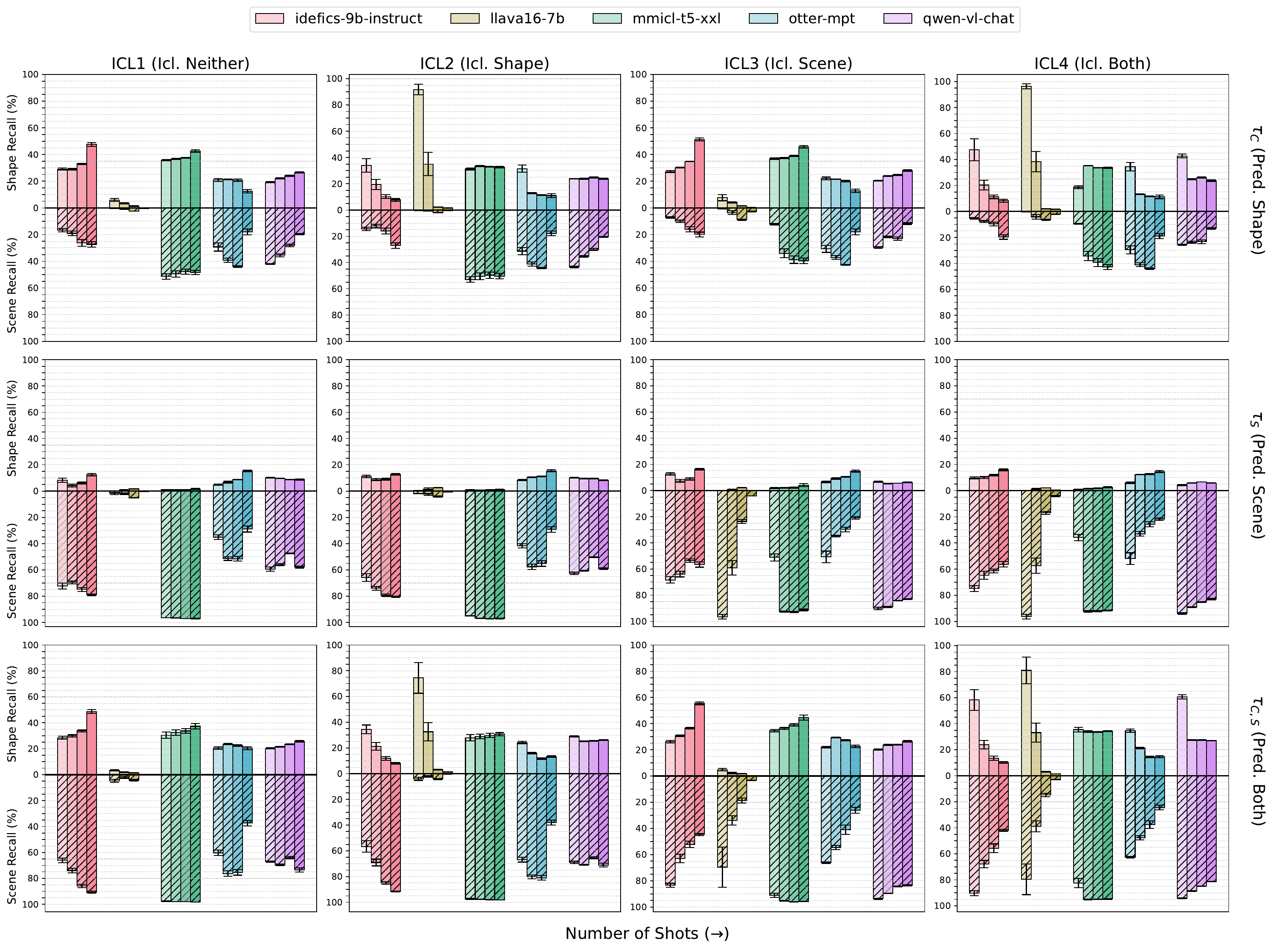}
    \caption{\textbf{Prompt sensitivity on \logos.} Mean shape and scene recall metrics with error bars representing one standard error from the mean across four different prompts used for ICL on the \logos dataset split.}
    \label{fig:propmt senstivity}
\end{figure}

We report the mean shape and scene recall on the \logos dataset split, with error bars representing one standard error from the mean. The results are shown in \Cref{fig:propmt senstivity}. Overall, we observe very little variation in shape and scene recall across models, tasks, and contexts. Significant variations, when present, occur only for LLava or Idefics models and are limited to cases with a small number of shots. These variations diminish as the number of in-context examples increases, suggesting that the results described in \Cref{sec:icl_exp} are generally insensitive to the type of prompt used, particularly when a larger number of in-context examples are provided.

\begin{figure}
    \centering
    \includegraphics[width=1.0\linewidth]{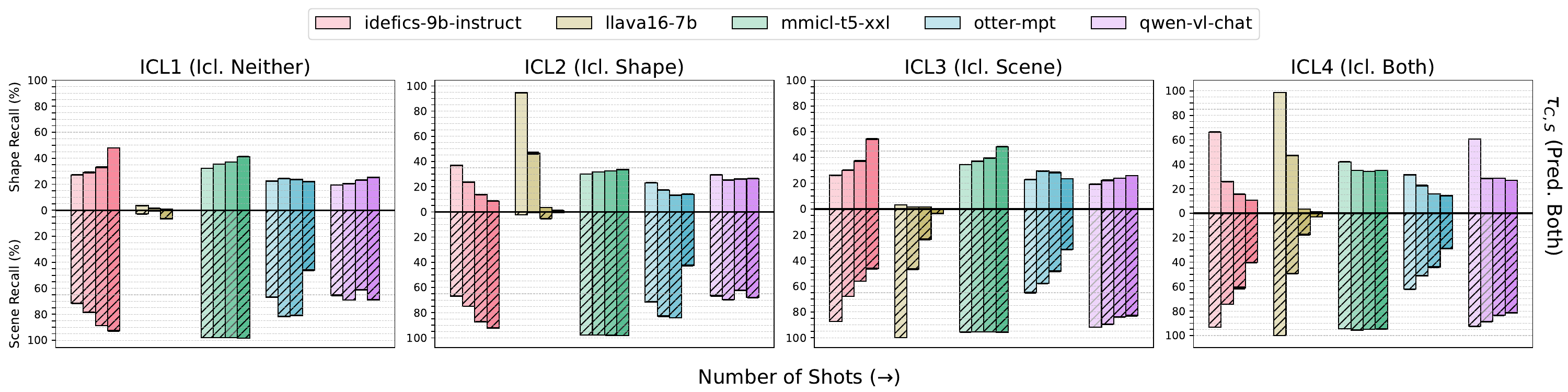}
    \caption{\textbf{Context ordering sensitivity on \logos.} Mean shape and scene recall metrics with error bars representing one standard error from the mean over three shuffled orders of the same context examples used for ICL on the \logos dataset split.}
    \label{fig:shuffle results}
\end{figure}

\textbf{Sensitivity to the order of in-context examples.} We also investigate the sensitivity of ICL results to the order of in-context examples. To assess this, we shuffle the context examples three times on the \logos when performing inference on task $\tau_{C,S}$. The results, shown in \Cref{fig:shuffle results}, display very tight metrics with minimal variation in shape and scene recall, demonstrating that the results described in \Cref{sec:icl_exp} are not sensitive to the ordering of the context examples.

\section{Domain Generalisation Experiments Details}
\label{app_mdg_full}
\subsection{Background Details}
\label{app:mdg_lit}
Domain generalisation has been a challenging task for image recognition. Several methods have been developed to improve training strategies for better generalisability of early specialist visual models, which are also applicable to CLIP models. Data augmentation strategies such as MixUp \cite{yan2020improve} and RegMixUp \cite{regmixup} are known to improve generalisation capacity through interpolation or extrapolation of data samples outside the training domain for diversity. GroupDRO \cite{sagawa2019distributionally} performs ERM with a re-weighting of classes with larger errors, making them more significant. VREx \cite{krueger2021out} reduces differences in risk across training domains, which can decrease a model’s sensitivity. Additionally, prompt learning, a promising approach for CLIP-style models, can also be leveraged for domain generalisation. Specifically, we adopt DPLCLIP \cite{zhang2023domain}, which trains a prompt generator during the training phase and infers unseen domains.

\subsection{Further Experiment Details}
\label{app:mdg_hyper}

\paragraph{CLIP Model} For all experiments, the image encoder backbone of CLIP model is a ResNet50 \cite{he2016deep}. For full-parameter fine-tuning, we train the whole image encoder, whereas for linear probing we only train the projection layer. The inferent prompt template for all methods is \texttt{``A photo of [Class name]''}.

\paragraph{Training Hyperparameters}
For all experiments, we use a batch size of 32 and the Adam optimiser \cite{kingma2014adam} with a learning rate of 5e-5. For full parameter fine-tuning, we train the model for 1000 steps, and for linear probing, we train the model for 800 steps. For MixUp \cite{yan2020improve} and RegMixUp \cite{regmixup}, the alpha and beta are both set to 0.2. For GroupDRO \cite{sagawa2019distributionally}, the eta is set to 1e-2. For VREx \cite{krueger2021out}, the penalty weight is set to 1.0. For DPLCLIP \cite{zhang2023domain}, the number of domain tokens is 16.

\section{Compute Resources}\label{app:misc}

All experiments are performed on our internal cluster. 

\paragraph{Resources for image generation}
For the Image generation, we used three A40 GPUs with 45 GB RAM with around 65h to generate all of the images in the dataset.

\paragraph{Resources for zero-shot experiments}
For the zero-shot experiments, we used eight A40 GPUs with 45 GB RAM for around 250h total to cover all Zero-shot experiments experiments.

\paragraph{Resources for in-context learning experiments} We perform ICL inference using 8 A40 GPUs with 45GB RAM for around 168h total to cover all ICL experimental settings.

\paragraph{Resources for domain generalisation experiments} For each fine-tuning CLIP we use a single A40 GPUs with 45GB RAM for an hour on average for full parameter fine-tuning and half an hour for linear probing.

%%%%%%%%%%%%%%%%%%%%%%%%%%%%%%%%%%%%%%%%%%%%%%%%%%%%%%%%%%%%

\end{document}